\documentclass{article} 
\usepackage{iclr2027_conference,times}

\usepackage{amsmath, mathtools, amssymb}
\usepackage{graphicx}
\usepackage{caption}   
\usepackage{subcaption}
\usepackage{multirow}
\usepackage{float}
\usepackage{hyperref}
\usepackage{booktabs}     
\usepackage{array}         
\usepackage{tabularx} 
\usepackage{blindtext}      
\usepackage{lipsum}         
\usepackage{titlesec}       
\usepackage{indentfirst}    
\usepackage{booktabs}       
\usepackage{array}               
\usepackage{tikz}          
\usepackage{pgfplots}
\usepackage{algorithm}
\usepackage{algpseudocode}
\usepackage{threeparttable}
\usepackage{xcolor}
\usepackage{url}
\usepackage{wrapfig}
\newcommand{\head}[1]{\paragraph{#1}}

\newcommand{\methodname}{{\textsc{Metta}}}

\title{
\methodname{}: Memory-Enhanced Test-Time\\
Adaptation for Meta-Reinforcement Learning
}

\author{
\textbf{Ali Najar \qquad Hosein Hasani \qquad Amir Kooshan Fattah}
\\[4pt]
\textbf{Yasamin Rajabi \qquad Mahdieh Soleymani Baghshah}
\\[6pt]
Department of Computer Engineering \\
Sharif University of Technology \\
\texttt{\{ali.najjar82,hosein.hasani,amir.hesari82,} \\
\texttt{Yasamin.rajabi53,soleymani\}@sharif.edu}
}

\iclrpreprintcopy
\begin{document}

\maketitle

\begin{abstract}
Meta-reinforcement learning aims to train agents that adapt quickly to unseen tasks from limited interaction, but this requires representing task evidence over long interaction histories. Classical recurrent agents compress experience into hidden states, while transformer-based agents retrieve from past trajectories at quadratic memory and computation cost. In this work, we propose Memory-Enhanced Test-Time Adaptation (\methodname{}) for meta-reinforcement learning. It decomposes the fast adaptation phase into two temporal scales: a causal neural memory performs fine-grained within-episode adaptation with linear sequence complexity, while compact episode-level summaries support long-range cross-episode task inference without attending over the full interaction history. Moreover, equipping agents with neural memory provides sufficient expressive capacity and test-time training capability, enabling effective adaptation to unseen held-out tasks and task families at meta-test time. On unseen held-out Meta-World ML10/45 test environments, \methodname{} achieves a success rate over $50\%$ while maintaining a low training budget and short wall-clock time. On long-context sparse-reward PointMaze tasks, it outperforms transformer-based baselines by exhibiting more structured exploration behavior. These results show that \methodname{} provides an efficient path toward more adaptive and scalable long-context meta-RL.
\end{abstract}

\section{Introduction}

Reinforcement learning (RL) studies how agents learn to make sequential decisions through interaction with an environment, using rewards as feedback to improve future behavior~\citep{sutton2018reinforcement}. Despite strong performance in fixed environments, standard RL methods often require large amounts of data to adapt to new tasks and generalize poorly under environment shifts~\citep{packer2018assessing, cobbe19generalization}. Meta-reinforcement learning addresses this limitation by training agents over a distribution of tasks so they can use past training experience to adapt quickly to unseen tasks from limited interaction at meta-test time~\citep{duan2016rl2,wang2016learning,finn2017maml,rakelly2019pearl}. 
A fundamental problem in meta-RL is representing task evidence collected during interaction with the environment~\citep{humplik2019taskinference,zintgraf2021varibad}. Black-box sequence-based agents learn this representation directly from trajectories using internal dynamics~\citep{duan2016rl2,wang2016learning,mishra2018snail,alver2021inside}. Latent-context methods instead make task inference more explicit by inferring a task variable or belief from collected experience~\citep{humplik2019taskinference,rakelly2019pearl,zintgraf2021varibad}.

Yet despite their differences, most approaches still rely on compressed task representations, which may restrict their capacity in meta-RL settings where task evidence may be sparse, delayed, or unevenly distributed across episodes.
Once experience is compressed into a latent task representation, the policy cannot easily recover specific past evidence that later becomes important.

\begin{figure}[t]
\centering
\includegraphics[width=0.99\linewidth]{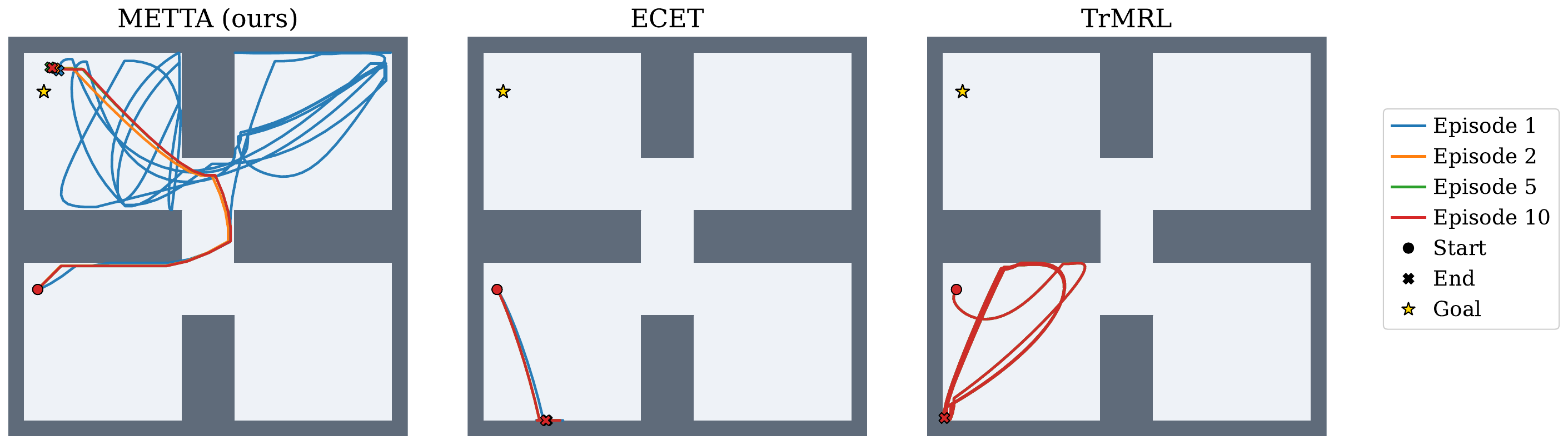}
\caption{Exploration behavior on a sparse-reward PointMaze task~\citep{pointmaze} across adaptation episodes. Trajectories are shown for Episodes $1$, $2$, $5$, and $10$ of the same task. As interaction proceeds, \methodname{} explores new regions and reduces repeated trajectories across episodes, suggesting more effective reuse of past exploration evidence in long-context adaptation. TrMRL~\citep{melo2022transformers} is difficult to scale to long multi-episode settings, while ECET~\citep{shala2025efficient} is limited by its short within-episode context.}
\label{fig:pointmaze_motivation}
\end{figure}

Transformers offer a natural alternative because self-attention enables associative retrieval over past interactions and supports long-range dependency modeling without relying on recurrent state compression~\citep{vaswani2017attention,melo2022transformers,shala2025efficient}. This makes them well suited to cross-episode meta-RL, where useful evidence may appear far back in the trial~\citep{mishra2018snail, shala2025efficient, melo2022transformers}. However, standard self-attention scales quadratically with sequence length~\citep{vaswani2017attention,duman2023computational}, making long adaptation horizons increasingly expensive in online RL settings. Prior work also showed that transformer optimization in reinforcement learning can be unstable due to exploration dynamics and non-stationary training signals~\citep{parisotto2019stabilizing}.
In online meta-RL, the effective context grows with both the episode horizon and the number of adaptation episodes, making full-history attention increasingly expensive in memory and computation. As a result, transformer-based agents often face a trade-off between retaining long histories and remaining computationally practical.

The computational overhead of full-history attention has motivated a recent shift toward modern linear recurrent models and neural memory architectures~\citep{gu2021s4,smith2022s5,gu2023mamba,sun2023retnet,sun2025ttt}. Instead of attending over the full history, these methods update an adaptive memory state online while preserving linear sequence complexity.
Recent work suggests that recurrence need not be limited to a small passive hidden state and can instead be equipped with expressive state dynamics or trainable memory modules, allowing models to store information over long horizons while preserving linear sequence complexity~\citep{gu2023mamba,sun2025ttt}. In particular, Test-Time Training (TTT)~\citep{sun2025ttt} treats memory as a trainable module that can be updated continually during inference, turning sequence processing from passive state compression into adaptive memory updating.
This has led to promising results in settings where long-range information is essential, including language modeling, time series, genomics, novel-view synthesis, and video generation~\citep{sun2025ttt,behrouz2025titans,zhang2025tttdoneright,dalal2025oneminute}.

In this work, we propose Memory-Enhanced Test-Time Adaptation (\methodname{}), which leverages TTT layers~\citep{sun2025ttt} as a modern neural memory module to gather task evidence from long-range interaction sequences in meta-RL. Besides this within-episode memory module, which provides fine-grained episode-level memories, we also use a coarse-grained aggregation mechanism that combines summaries across episodes.
Thus, in addition to the classical slow and fast learning processes in meta-RL, our fast adaptation phase separates within-episode adaptation from cross-episode evidence aggregation, allowing the policy to condition on longer meta-test histories without relying on quadratic computation over agent interaction steps.

Our experiments across locomotion, manipulation, and navigation benchmarks show that \methodname{} improves fast adaptation in both standard and long-context meta-RL settings. Interestingly, the relative gain over existing methods is especially notable when generalizing to unseen held-out Meta-World task families that are not encountered during meta-training. This provides evidence of task-family-level out-of-distribution generalization, where adaptation transfers to task families not observed during meta-training. This suggests that neural memory modules can support adaptation beyond the task families observed during training.

In addition, \methodname{} achieves these improvements with substantially lower training cost than transformer-based meta-RL methods, while preserving linear sequence complexity over long horizons.
Figure~\ref{fig:pointmaze_motivation} illustrates the role of neural memory in long-context sparse-reward settings. \methodname{} explores more broadly while reducing repeated trajectories across episodes, suggesting more effective reuse of information gathered during earlier episodes. We examine this behavior further in Section~\ref{sec:long_context_memory}.


\section{Preliminaries}

\subsection{Meta-Reinforcement Learning}

Reinforcement learning problems are formulated as Markov decision processes (MDPs), defined by the tuple $(\mathcal{S}, \mathcal{A}, P, R, P_0, \gamma, H)$, where $\mathcal{S}$ and $\mathcal{A}$ denote the state and action spaces, $P(s' \mid s,a)$ denotes the transition distribution, $R(s,a)$ the reward function, $P_0$ the initial-state distribution, $\gamma \in [0,1]$ the discount factor, and $H$ the horizon.
At timestep $t$, the agent observes state $s_t$, samples an action $a_t \sim \pi_\psi(\cdot \mid s_t)$ from a policy parameterized by $\psi$, receives reward $r_t$, and transitions to the next state $s_{t+1}$. The objective is to maximize the expected discounted return
\begin{equation}
J(\pi_\psi)
=
\mathbb{E}_{\pi_\psi}
\left[
\sum_{t=0}^{H}
\gamma^t r_t
\right].
\end{equation}

In meta-reinforcement learning, tasks are sampled from a distribution $p(\mathcal{T})$, where each task
$\mathcal{T}_i = (\mathcal{S}, \mathcal{A}, P_i, R_i, P_{0,i}, \gamma, H)$
shares the same observation and action spaces but may differ in transition dynamics, reward functions, or goals.
During meta-training, the agent interacts with tasks sampled from $p(\mathcal{T})$ and learns a policy that can adapt efficiently across the task distribution. The meta-RL objective is to maximize expected return across the task distribution,
\begin{equation}
\max_\psi
\;
\mathbb{E}_{\mathcal{T}\sim p(\mathcal{T})}
\left[
\mathbb{E}_{\pi_\psi}
\left[
\sum_{t=0}^{H}
\gamma^t r_t
\right]
\right].
\label{eq:objective}
\end{equation}

During meta-testing, the learned policy is evaluated on unknown tasks sampled from the same distribution $p(\mathcal{T})$.
Meta-RL evaluation commonly measures adaptation over multiple episodes from the same task instance or family. The agent must improve behavior using only its internal memory and interaction history during rollout execution.

\subsection{Neural Memory}

Neural memory provides a common abstraction for sequence models that update an internal state and read from it to produce predictions. This view includes recurrent networks, linear attention, state-space models, and recent test-time memory architectures~\citep{katharopoulos2020transformers,gu2023mamba,sun2025ttt,behrouz2025titans}. Given an input sequence $x_{1:n}$, a memory model maintains a state $\mathcal{M}_t=\mathcal{U}_\theta(\mathcal{M}_{t-1},x_t)$ and produces an output $y_t=\mathcal{R}_\theta(\mathcal{M}_t,x_t)$, where $\mathcal{U}_\theta$ is the update rule and $\mathcal{R}_\theta$ is the readout rule. This notation covers both fixed-size recurrent states and richer memory states designed to store and retrieve information over longer horizons.




Many memory models use an attention-style interface in which each input is projected into a key, value, and query. Given learned projections $\mathbf{k}_t=\theta_K x_t$, $\mathbf{v}_t=\theta_V x_t$, and $\mathbf{q}_t=\theta_Q x_t$, the memory update and readout can be written in terms of key-value writing and query-based retrieval.
In general, the write and read operations can be expressed as
\[
    \mathcal{M}_t
    =
    f\!\left(\mathcal{M}_{t-1}, \mathbf{k}_t, \mathbf{v}_t\right),
    \qquad
    y_t
    =
    g\!\left(\mathcal{M}_t, \mathbf{q}_t\right),
\]

where $f$ is the memory update rule, and $g$ is the read function. A simple example is the linear Transformer~\citep{katharopoulos2020transformers,schlag2021linear}, which can be interpreted as a fast-weight memory storing key-value associations. In this case 
$
    f\!\left(\mathcal{M}_{t-1}, \mathbf{k}_t, \mathbf{v}_t\right)
    =
    \mathcal{M}_{t-1}
    +
    \mathbf{v}_t \phi(\mathbf{k}_t)^\top
$
and 
$ g\!\left(\mathcal{M}_t, \mathbf{q}_t\right) = \mathcal{M}_t \phi(\mathbf{q}_t) / \mathbf{n}_t^\top \phi(\mathbf{q}_t), 
$
where $\phi(\cdot)$ is a feature map and $\mathbf{n}_t = \sum_{i=1}^{t} \phi(\mathbf{k}_i)$ is the normalization state. This example shows how attention-like retrieval can be implemented through a recurrent memory update.

\paragraph{Test-Time Training (TTT).}
One recent neural memory model with linear sequence complexity is TTT, whose hidden states are themselves learnable models updated while processing the sequence~\citep{sun2025ttt}. Given an input sequence $x_{1:n}$, a TTT layer maintains an inner-loop model $f_{\mathcal{M}_t}$, where $\mathcal{M}_t$ denotes the hidden state after processing the prefix $x_{1:t}$. The layer therefore keeps the interface of a recurrent sequence model, but replaces the usual hidden vector with the parameters of a model that is trained online.

The memory is updated by training $f_\mathcal{M}$ in a self-supervised manner to predict the label view $\mathbf{v}_t$ from the training view $\mathbf{k}_t$,
\begin{align}
    \ell(\mathcal{M}; \mathbf{k}, \mathbf{v})
    =
    \frac{1}{2}
    \left\|
        f_{\mathcal{M}}(\mathbf{k}) - \mathbf{v}
    \right\|_2^2,
    \qquad
    \mathcal{M}_t
    =
    \mathcal{M}_{t-1}
    -
    \eta_t
    \nabla_{\mathcal{M}}
    \ell(\mathcal{M}_{t-1}; \mathbf{k}_t, \mathbf{v}_t),
    \label{eq:selfsup_loss}
\end{align}

where $\eta_t$ is a learnable inner-loop learning rate. The output is then produced by applying the updated memory to the test view,
$
    y_t = f_{\mathcal{M}_t}(\mathbf{q}_t).
$
Thus, $y_t$ is computed from a memory model that has been adapted on the causal prefix $x_{1:t}$.


The distinction between inner-loop and outer-loop parameters is central to TTT. The memory parameters $\mathcal{M}_t$ are \emph{fast} parameters: they are initialized at the start of a sequence and updated while the sequence is processed, including at test time. In contrast, the view projections $\theta_K,\theta_V,\theta_Q$, the initial memory parameters, and the downstream network parameters are \emph{slow} parameters learned across training sequences. Thus, the outer loop learns how to construct a useful self-supervised update rule, while the inner loop applies this rule online to adapt the memory to the current sequence.

\section{\methodname{}}

In this section, we introduce Memory-Enhanced Test-Time Adaptation (\methodname{}) for meta-RL.
We consider a trial-based meta-RL setting in which the agent interacts with multiple consecutive episodes from the same sampled task. Each trial is generated from a hidden task $\mathcal{T} \sim p(\mathcal{T})$ and contains $E$ episodes, each with horizon $H$.

At episode $e$ and timestep $t$, the agent uses the interaction history to construct a context representation $z_{e,t}$ that summarizes task-relevant evidence accumulated during adaptation. The policy conditions on both the current state and this context representation for action selection,
\begin{equation}
a_{e,t}
\sim
\pi_\psi
\left(
\cdot
\mid
s_{e,t}, z_{e,t}
\right).
\label{eq:policy}
\end{equation}

\begin{figure}[t]
\centering
    \includegraphics[width=0.95\linewidth]{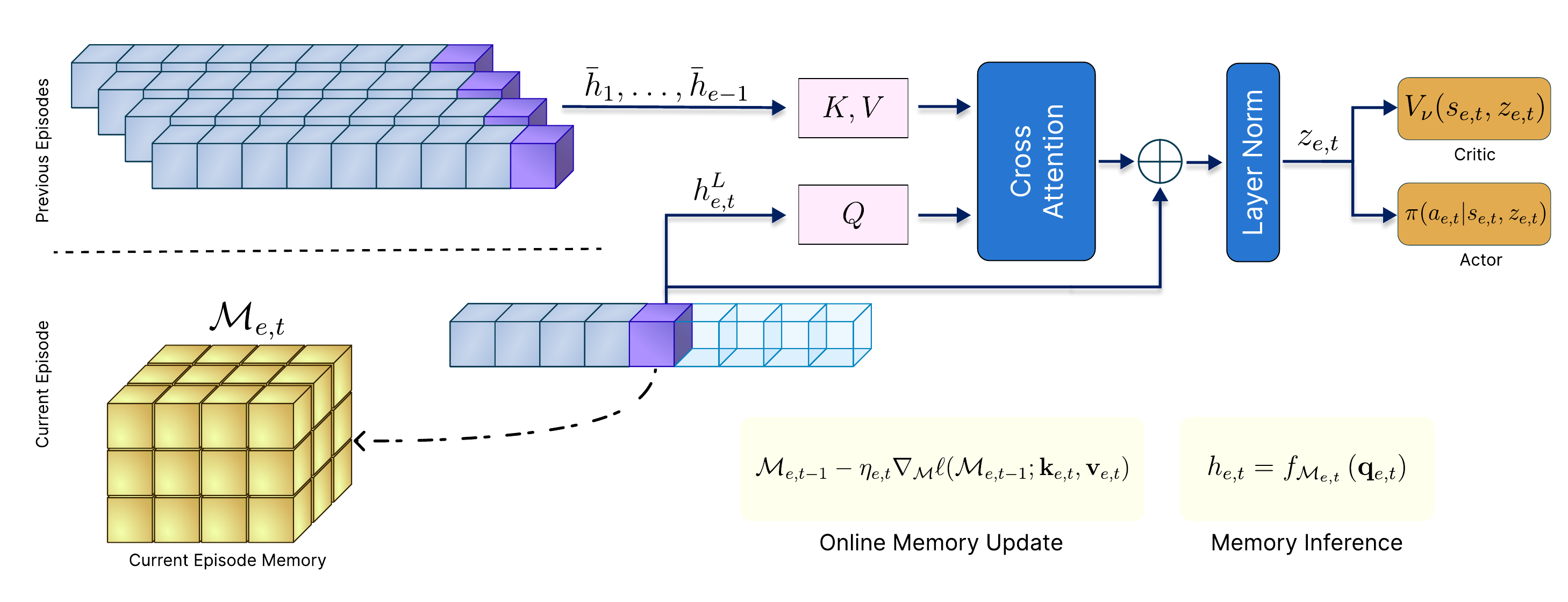}
    \caption{
    Overview of \methodname{}. 
    The current episode is processed by a within-episode neural memory to produce $h_{e,t}^{(L)}$. At the end of each episode, the final representation is stored as an episode summary $\bar{h}_e$. During later episodes, previous summaries are
    combined with the current representation to form the context representation $z_{e,t}$. The actor and critic condition on the current state $s_{e,t}$ and $z_{e,t}$.
    }
\label{fig:main}
\end{figure}

Besides the standard slow parameter updates during meta-training, \methodname{} performs fast adaptation at two temporal scales. A within-episode neural memory supports fine-grained online adaptation, while a cross-episode memory aggregates information across episodes. Together, these modules produce the context representation $z_{e,t}$, which provides the policy with information accumulated through interaction. Figure~\ref{fig:main} summarizes the overall architecture.

All modules are trained end-to-end with PPO~\citep{schulman2017ppo} over full meta-RL trials. Additional implementation details and full meta-train and meta-test algorithms are provided in Appendix~\ref{app:implement}.

\subsection{Within-Episode Neural Memory}

We use a TTT architecture~\citep{sun2025ttt} as the within-episode neural memory in \methodname{}. We denote this module by $\mathrm{TTT}_\theta$, which consists of $L$ stacked TTT layers whose fast weights are updated online while processing each token. METTA is modular with respect to the within-episode memory architecture. We instantiate this component with TTT in our default model, but the framework can accommodate other causal neural-memory architectures, allowing us to separately study the role of memory design and cross-episode aggregation.

Following RL$^2$-style conditioning~\citep{duan2016rl2}, \methodname{} processes the sequential input
$x_{e,t} = [s_{e,t}, a_{e,t-1}, r_{e,t-1}, d_{e,t-1}]$
in an autoregressive manner. 
Sequence processing is performed by $\mathrm{TTT}_\theta$ with linear computational cost. We first embed the interaction input as $h_{e,t}^{(0)}=\phi(x_{e,t})$, and then pass this representation through $L$ TTT layers. For each layer $l$, the model maintains a fast-weight memory $\mathcal{M}_{e,t}^{(l)}$ that is initialized at the start of the episode and updated online as the episode unfolds.

Given the previous-layer representation $h_{e,t}^{(l-1)}$, the layer forms training, label, and test views,
\[
    \mathbf{k}_{e,t}^{(l)}
    =
    \theta_K^{(l)} h_{e,t}^{(l-1)},
    \qquad
    \mathbf{v}_{e,t}^{(l)}
    =
    \theta_V^{(l)} h_{e,t}^{(l-1)},
    \qquad
    \mathbf{q}_{e,t}^{(l)}
    =
    \theta_Q^{(l)} h_{e,t}^{(l-1)}.
\]
It first writes the current evidence into memory through a gradient step on the self-supervised loss defined in Eq.~\eqref{eq:selfsup_loss},
\[
    \mathcal{M}_{e,t}^{(l)}
    =
    \mathcal{M}_{e,t-1}^{(l)}
    -
    \eta_{e,t}^{(l)}
    \nabla_{\mathcal{M}}
    \ell
    \left(
        \mathcal{M}_{e,t-1}^{(l)};
        \mathbf{k}_{e,t}^{(l)}, \mathbf{v}_{e,t}^{(l)}
    \right).
\]
The updated memory is then queried to produce the next-layer representation,
$h_{e,t}^{(l)} = f_{\mathcal{M}_{e,t}^{(l)}}(\mathbf{q}_{e,t}^{(l)})$.
After the final layer, $h_{e,t}^{(L)}$ is the adapted context
representation of the current episode prefix.

\subsection{Cross-Episode Memory}

The within-episode TTT module adapts to the current episode, but meta-RL trials also contain useful information across episodes. We use the final adapted representation of each episode, $\bar{h}_e = h_{e,H}^{(L)}$, as an episode-level summary. At episode $e$, the agent causally has access to all previous summaries $\mathcal{H}_{e-1}=\{\bar{h}_i\}_{i=1}^{e-1}$. We consider two mechanisms for
aggregating these summaries to construct the context representation $z_{e,t}$, which is used as input to the actor and critic heads.

\paragraph{Exponential moving average.}
A simple aggregation mechanism is to maintain an exponential moving average (EMA) of previous episode summaries. Let $c_{e-1}$ denote the compact cross-episode summary available before episode $e$. 
For $e>1$, the current within-episode representation is combined with the recent cross-episode summary to form the current context representation
\begin{equation}
    z_{e,t}
    =
    \alpha c_{e-1}
    +
    (1-\alpha)h_{e,t}^{(L)},
    \label{eq:ema_read}
\end{equation}
where $\alpha \in [0,1]$ controls the influence of previous episodes. For the first episode, no previous summary is available, so $z_{1,t}=h_{1,t}^{(L)}$ and $c_1 = \bar{h}_1$. The EMA summary for episode $e$ is updated as
\begin{equation}
    c_e = \alpha c_{e-1} + (1-\alpha)\bar{h}_e.
    \label{eq:ema_memory}
\end{equation}
This variant stores only a single vector and updates cross-episode information in constant time.

\paragraph{Attention-based retrieval.}
While EMA is efficient, it does not allow selective retrieval of previous episodes. Therefore, we use an attention layer over stored episode summaries as the default aggregation mechanism in \methodname{}. This learnable attention layer forms a query $q_{e,t}=W_q h_{e,t}^{(L)}$ and uses the unordered set of episode summaries $\mathcal{H}_{e-1}$ as keys and values for information retrieval:
\begin{equation}
    \tilde{h}_{e,t}
    =
    \mathrm{Attn}
    \left(
        q_{e,t}, \mathcal{H}_{e-1}
    \right).
    \label{eq:memory_attention}
\end{equation}
The retrieved cross-episode representation is fused with the current within-episode representation,
\begin{equation}
    z_{e,t}
    =
    \mathrm{LayerNorm}
    \left(
        h_{e,t}^{(L)} + \tilde{h}_{e,t}
    \right).
    \label{eq:memory_fusion}
\end{equation}
The resulting representation $z_{e,t}$ is passed to the RL agent (actor and critic modules) along with the current state $s_{e,t}$. Compared with EMA, attention-based retrieval is more expressive because it can learn which previous episode summaries contain the most relevant task information.

\section{Experiments}

\subsection{Experimental Setup}

\paragraph{Benchmarks.}
We evaluate the effectiveness of our method on standard meta-RL benchmarks covering locomotion, manipulation, and navigation tasks.
We evaluate continuous-control adaptation on \textbf{MuJoCo} HalfCheetahVel~\citep{mujoco}, where the agent must infer the target velocity from interaction. We use a limited training budget to compare how efficiently each method adapts under restricted environment interaction.

Next, we evaluate robotic manipulation on ML1, ML10, and ML45 from the \textbf{Meta-World} benchmark~\citep{metaworld}. In ML1, following the standard baseline convention, goal coordinates are masked and must be inferred through interaction. ML10 and ML45 span multiple task families with disjoint train and test splits, allowing us to evaluate generalization to families not observed during meta-training. We view this as a form of task-family-level OOD evaluation, since the held-out test families are encountered only at meta-test time. In these settings, goal coordinates are included in the observation, while the task family remains fixed within each trial. During multi-class training, each rollout batch assigns environments to task classes in a balanced way and then samples a task variation within each class.

Finally, we use \textbf{PointMaze}~\citep{pointmaze} with sparse rewards (only at the goal position) to evaluate long-context adaptation. In this environment, the agent may need to explore for many steps before receiving useful feedback, making it important to preserve information from earlier parts of the trajectory. This benchmark directly tests whether meta-RL methods can benefit from long context.

\begin{figure}[t]

\centering

\begin{subfigure}[t]{0.485\linewidth}
    \centering
    \begin{minipage}[t]{0.494\linewidth}
        \centering
        \includegraphics[width=\linewidth]{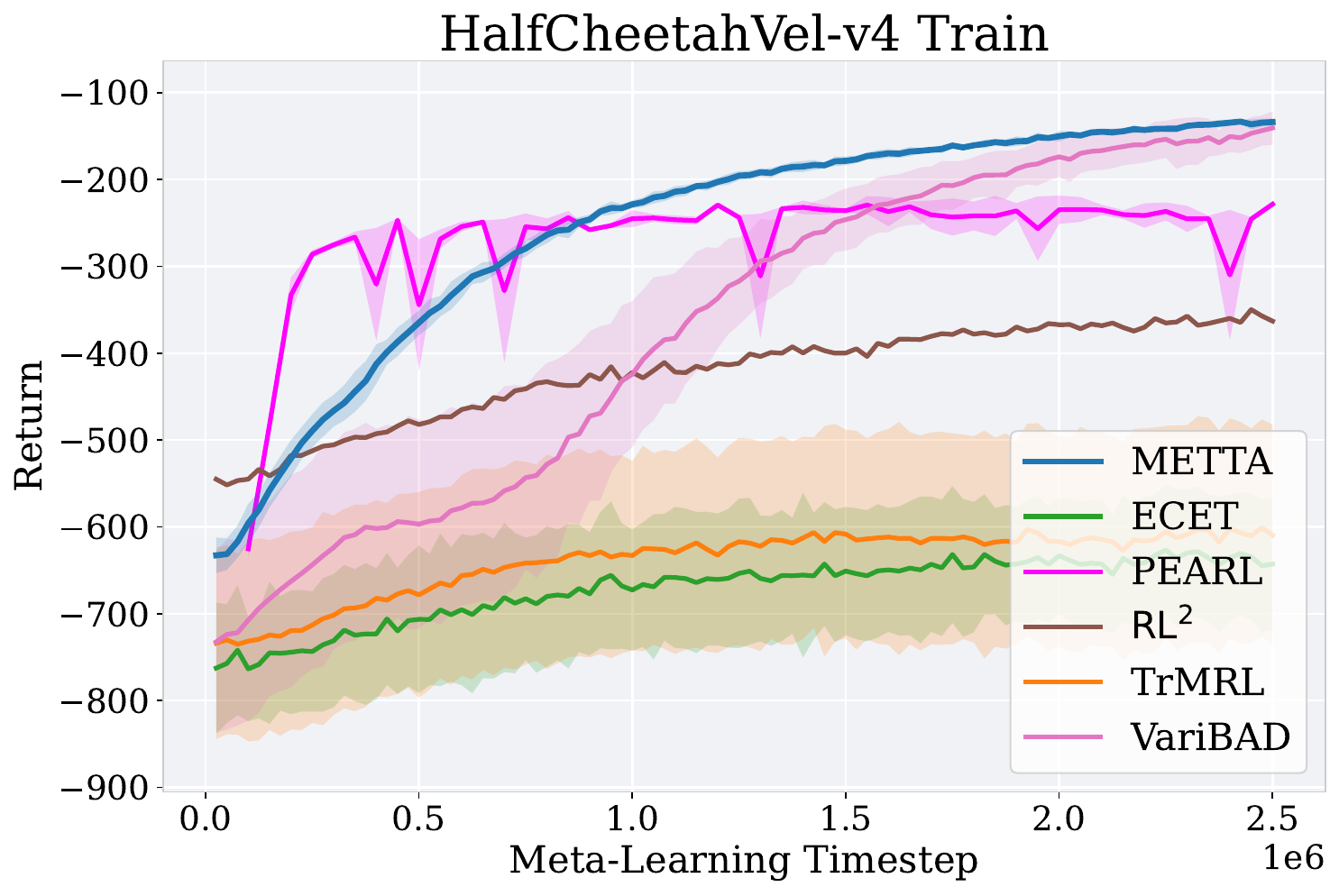}
    \end{minipage}
    \begin{minipage}[t]{0.494\linewidth}
        \centering
        \includegraphics[width=\linewidth]{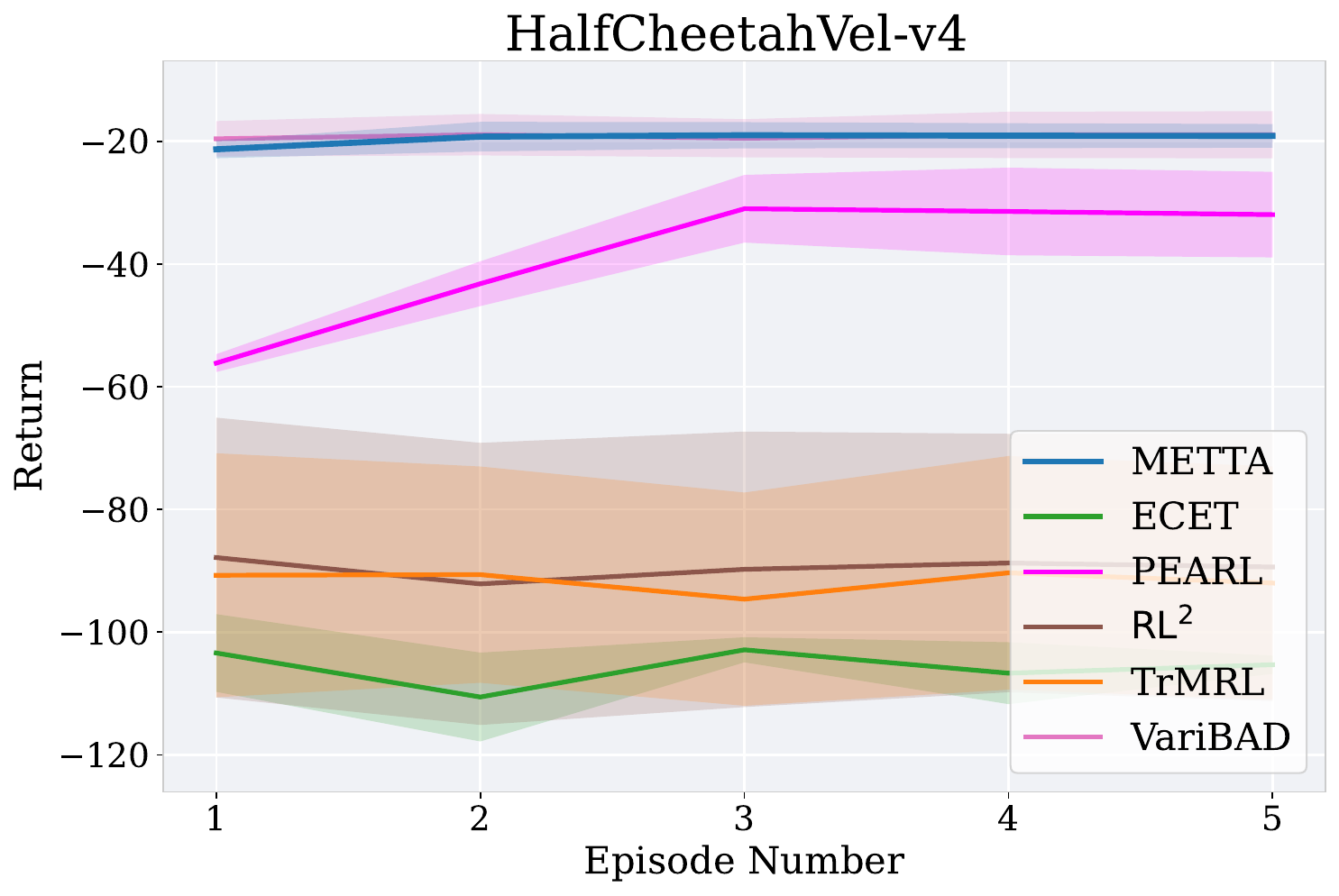}
    \end{minipage}
    \caption{MuJoCo HalfCheetahVel.}
    \label{fig:mujoco_results}
\end{subfigure}
\hfill
\begin{subfigure}[t]{0.485\linewidth}
    \centering
    \begin{minipage}[t]{0.494\linewidth}
        \centering
        \includegraphics[width=\linewidth]{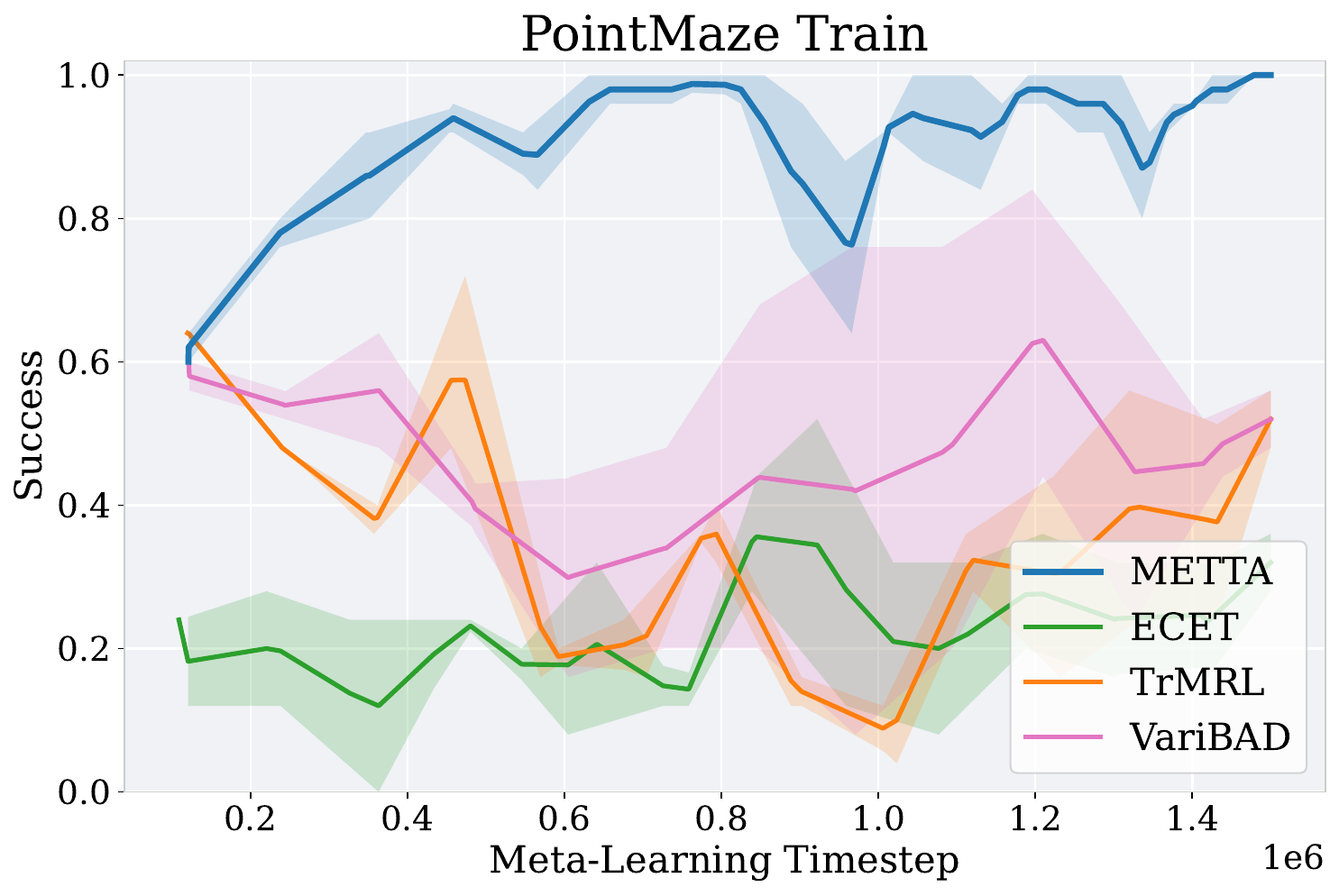}
    \end{minipage}
    \begin{minipage}[t]{0.494\linewidth}
        \centering
        \includegraphics[width=\linewidth]{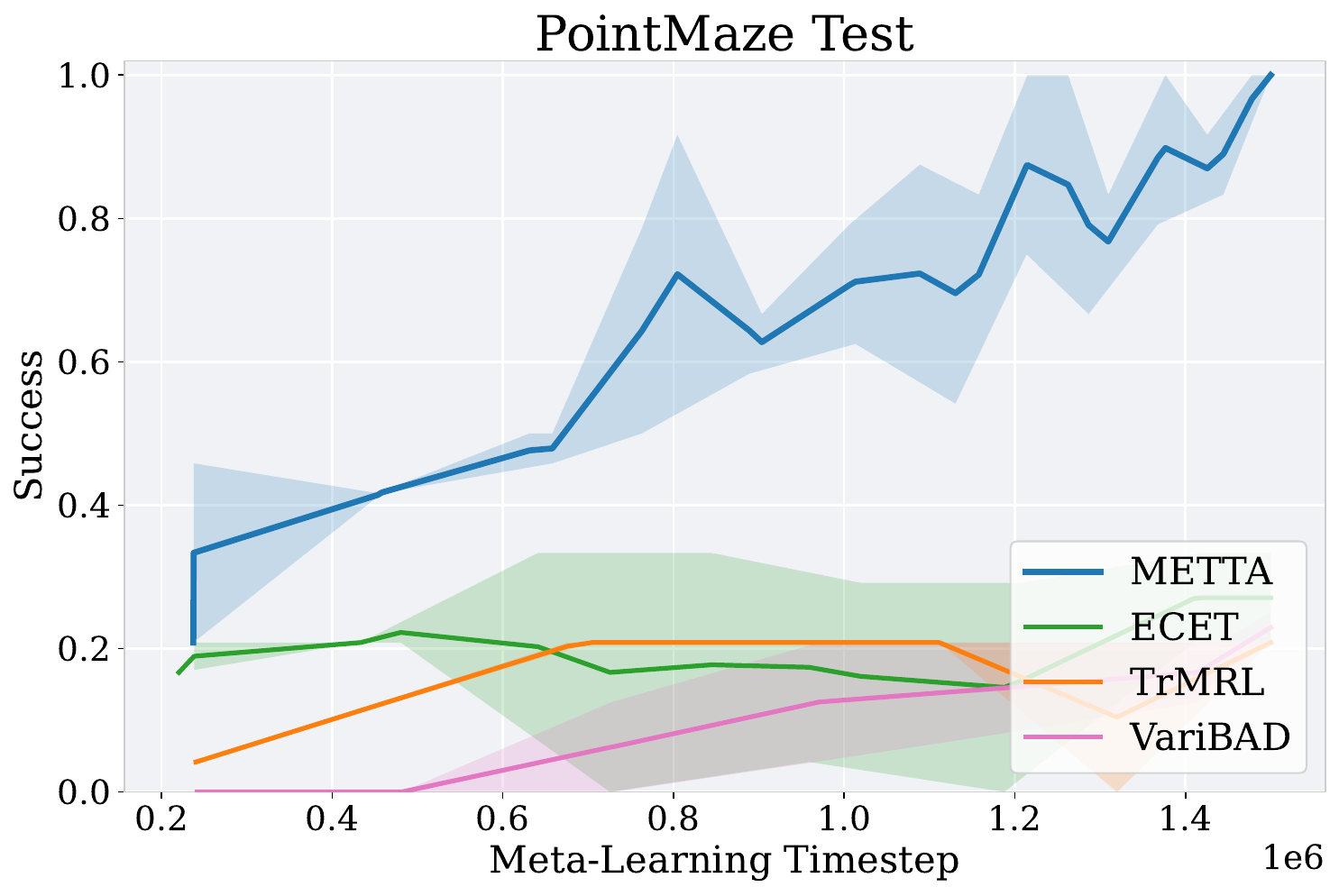}
    \end{minipage}
    \caption{Sparse-reward PointMaze.}
    \label{fig:pointmaze_results}
\end{subfigure}

\caption{
Fast adaptation during meta-training and meta-testing.
(a) HalfCheetahVel training return and adaptation across meta-test episodes.
(b) PointMaze training and unseen held-out test success rate.
}
\vspace{-3pt}
\label{fig:mujoco_pointmaze_results}
\end{figure}

\paragraph{Baselines.}
We compare \methodname{} against representative meta-RL baselines from recurrent, gradient-based, latent-context, and transformer-based families. RL$^2$ uses recurrent state to implement black-box adaptation from interaction history~\citep{duan2016rl2}, while MAML learns an initialization that can be adapted with gradient updates~\citep{finn2017maml}. PEARL and VariBAD represent latent-context approaches that infer task variables from collected experience~\citep{rakelly2019pearl,zintgraf2021varibad}. For transformer-based meta-RL, we compare against TrMRL, which uses self-attention over trajectory context~\citep{melo2022transformers}, ECET, which improves cross-episode conditioning~\citep{shala2025efficient}, and AMAGO, an in-context RL agent trained on long sequences~\citep{grigsby2024amago}.
For all baselines, we use the default implementations and hyperparameters from their respective repositories. All methods use the same evaluation protocol. Results are averaged over five
seeds, with shaded regions showing $\pm 1$ SEM. Additional details are provided in Appendix~\ref{app:experimental_protocol}.

\begin{figure}[t]
\centering
\includegraphics[width=\linewidth]{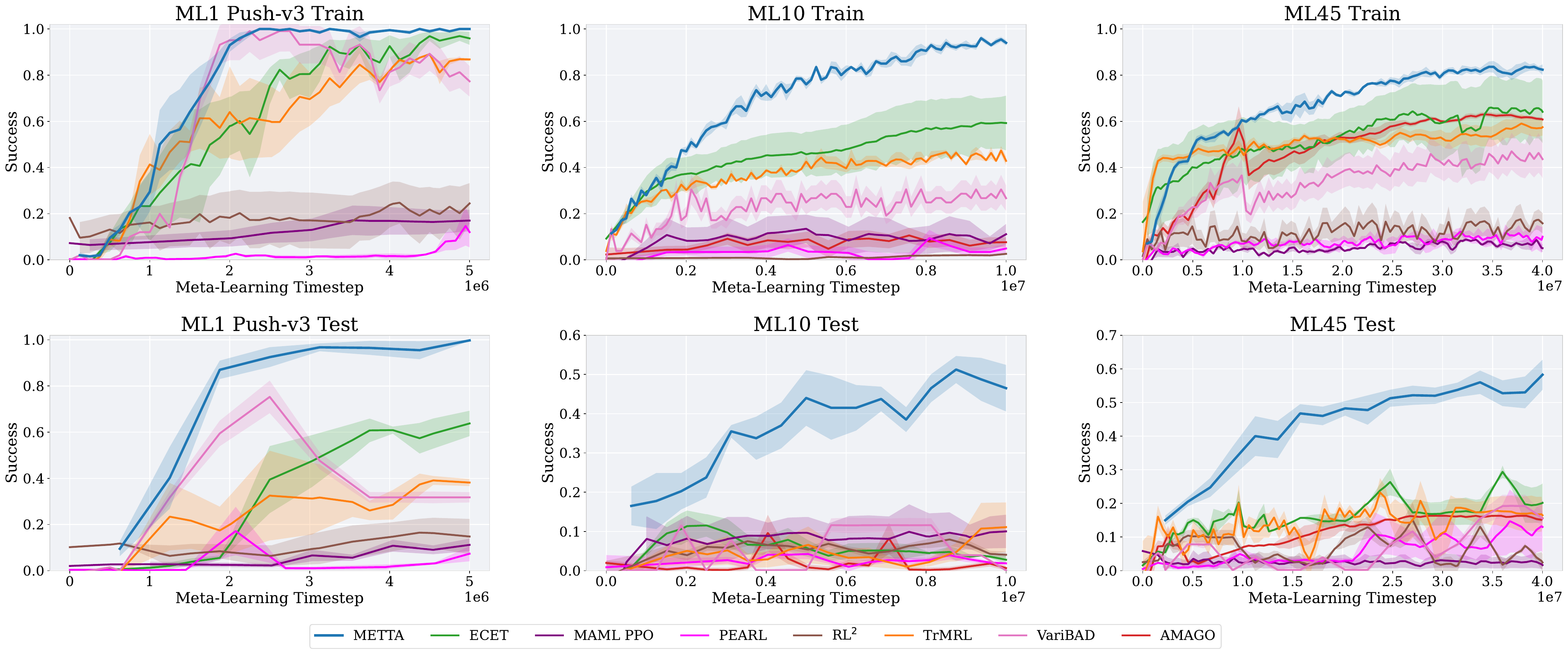}

\caption{
Meta-World success rate on ML1, ML10, and ML45. Top: evaluation on unseen task instances from task families used during meta-training. Bottom: evaluation on the Meta-World test benchmark, which contains unseen held-out task families for ML10 and ML45.
}
\vspace{-3pt}
\label{fig:metaworld_res}
\end{figure}

\subsection{Main Results}

In this section, we focus on standard evaluations of baselines using well-known metrics, average return and success rate, and leave additional analyses, including ablation studies and qualitative results, to Appendix~\ref{app:additional_res}.
Figure~\ref{fig:mujoco_results} reports HalfCheetahVel performance under a limited training budget. The training curve shows that \methodname{} improves rapidly with low training variance. The episode-wise adaptation plot in Figure~\ref{fig:mujoco_results} shows that \methodname{} maintains strong returns from the first episode and remains stable across later episodes, suggesting that its neural memory can quickly infer the target velocity from interaction.

\subsubsection{Role of Long-Context Memory} 
\label{sec:long_context_memory}

Figure~\ref{fig:pointmaze_results} shows that, in long-horizon tasks with sparse rewards, \methodname{} rapidly improves the success rate during both meta-training and meta-testing, while the other baselines remain substantially lower. This suggests that \methodname{} can preserve useful evidence from earlier exploration and reuse it in later episodes. Transformer-based baselines~\citep{melo2022transformers,shala2025efficient} struggle in this setting. TrMRL is difficult to scale to long effective horizons and tends to repeat similar exploration behavior, while ECET is limited by short-context cross-episode conditioning. VariBAD improves more slowly, likely because sparse rewards provide weak evidence for updating the latent task belief until successful exploration occurs~\citep{zintgraf2021varibad}. 


\begin{wraptable}{r}{0.58\textwidth}
\vspace{-10pt}
\centering
\caption{PointMaze exploration at 1.5M environment steps (mean $\pm$ SEM).}
\label{tab:pointmaze_exploration}
\small
\setlength{\tabcolsep}{3pt}
\begin{tabular}{lccc}
\toprule
Method & Coverage $\uparrow$ & Episode Jaccard $\downarrow$ & Steps $\downarrow$ \\
\midrule
\methodname{} & $\mathbf{0.447 \pm 0.036}$ & $\mathbf{0.792 \pm 0.022}$ & $\mathbf{194.8 \pm 30.1}$ \\
ECET & $0.250 \pm 0.087$ & $0.935 \pm 0.028$ & $427.3 \pm 44.5$ \\
TrMRL & $0.330 \pm 0.031$ & $1.000 \pm 0.000$ & $373.4 \pm 16.1$ \\
\bottomrule
\end{tabular}
\vspace{-8pt}
\end{wraptable}

The exploration patterns in Figure~\ref{fig:pointmaze_motivation} and the additional trajectories in Appendix~\ref{app:qualitative} are further supported by Table~\ref{tab:pointmaze_exploration}. Coverage measures the fraction of the traversable maze visited, while Episode Jaccard measures overlap between consecutive episodes, with lower values indicating less repetitive exploration. Steps measures the interactions required before success or horizon termination. \methodname{} achieves the highest coverage and lowest overlap while requiring the fewest steps, suggesting that it maintains diverse exploration across episodes without sacrificing navigation efficiency. In contrast, ECET and TrMRL cover less of the maze while revisiting previously explored regions more frequently. Detailed definitions are provided in Appendix~\ref{app:pointmaze_metrics}.


Figure~\ref{fig:metaworld_res} shows adaptation performance on the Meta-World benchmark. On simpler settings such as ML1, \methodname{} quickly improves over adaptation episodes, indicating that the neural memory can infer task-relevant information from limited interaction. On the more challenging ML10 and ML45 benchmarks, \methodname{} surpasses other baselines by a large margin across diverse task families.

The performance gap is notable not only on train environments, where evaluation uses task families observed during meta-training, but also on the Meta-World test environments (up to a $\sim40\%$ success-rate margin), which consist of unseen held-out task families. This result indicates that the learned adaptation mechanism of \methodname{} generalizes beyond the task families encountered during meta-training. In our experiments, this generalization is substantially stronger than that of RL$^2$~\citep{duan2016rl2}, suggesting an advantage of adaptive neural memory over conventional autoregressive recurrent conditioning, which relies on passive and less expressive memory updates during inference.

\begin{figure}[t]
\centering

\begin{minipage}[t]{0.49\linewidth}
    \vspace{2pt}
    \centering
    \includegraphics[width=\linewidth]{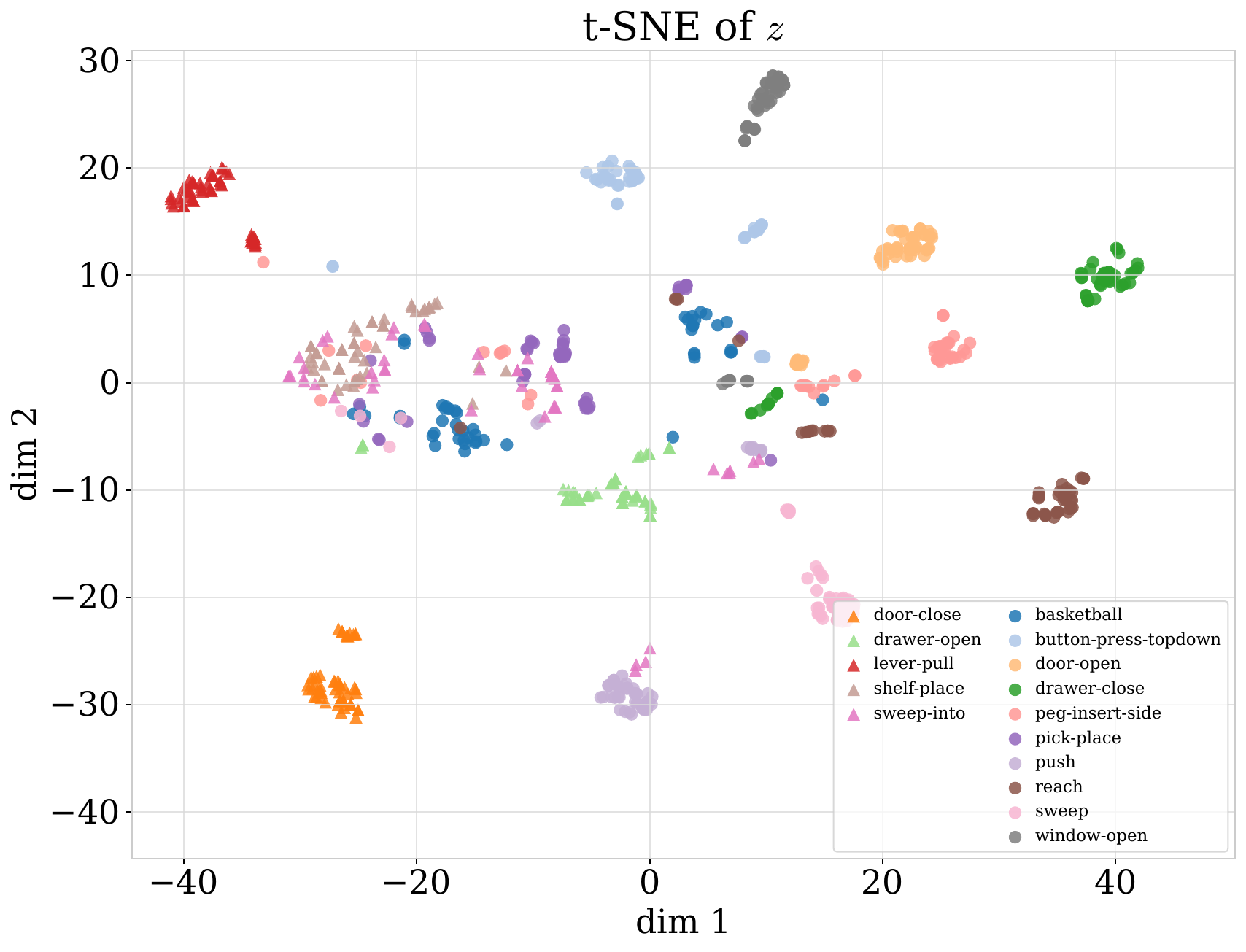}
\end{minipage}
\hspace{0.1cm}
\begin{minipage}[t]{0.44\linewidth}
    \vspace{0pt}
    \centering
    \includegraphics[width=\linewidth]{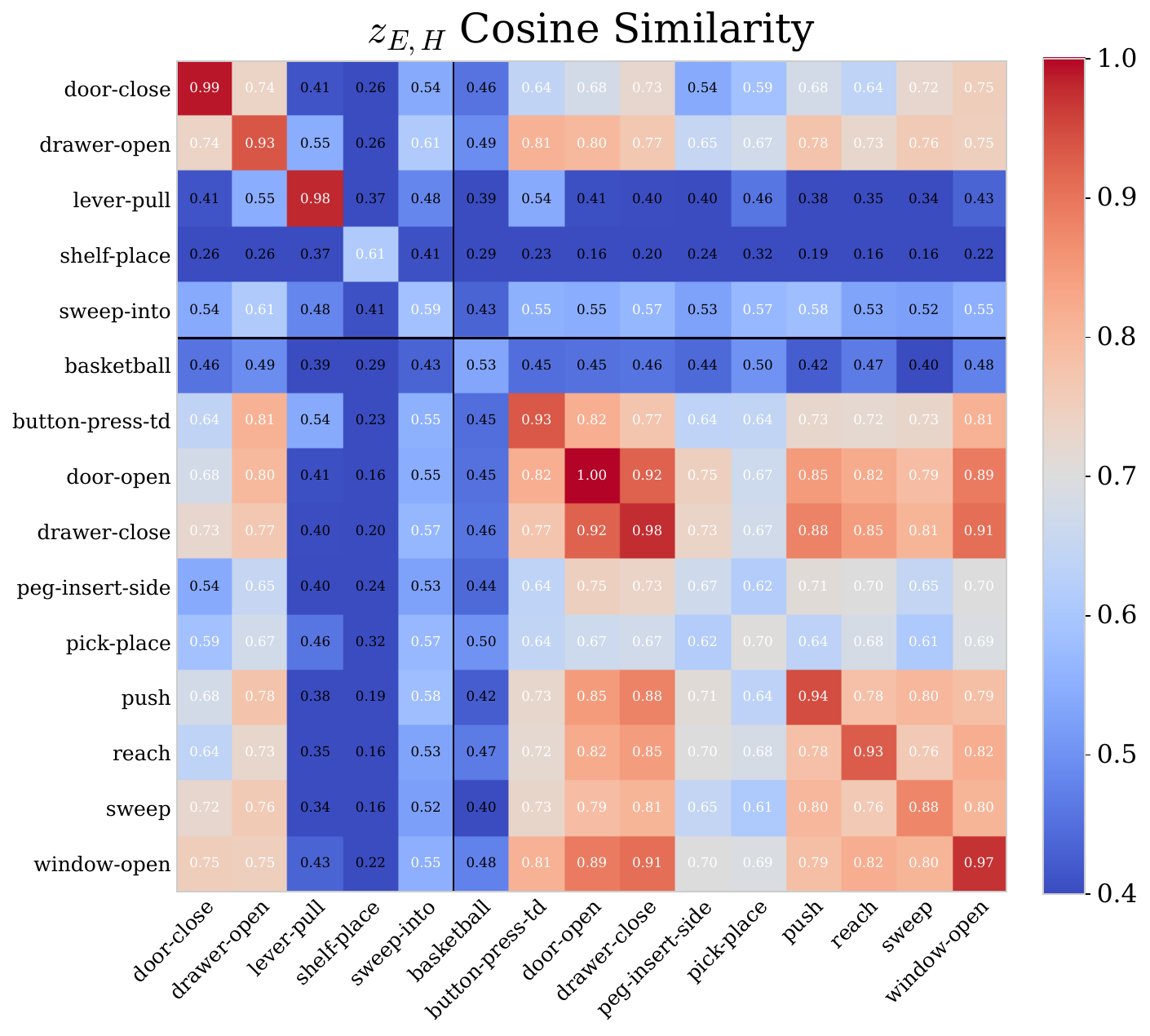}
\end{minipage}

\caption{
Structure of the learned context representations.
Left: t-SNE visualization of $z$ for all episodes. Circles denote training tasks and triangles denote test tasks. Right: pairwise cosine-similarity matrix computed from the final episode representation of each trial.
}
\vspace{-3pt}
\label{fig:representation_analysis}
\end{figure}

To further analyze the structure of the learned context, we measure representation similarity both quantitatively and qualitatively. Figure~\ref{fig:representation_analysis} shows cosine similarities of the context representations on ML10 along with a t-SNE visualization. The t-SNE plot shows that \methodname{} organizes context representations according to task families while keeping test tasks aligned with related training tasks, suggesting generalization beyond memorization. The cosine-similarity matrix shows consistent representations within each task family through large diagonal values, while the off-diagonal structure captures relationships across different task families. These results indicate that the learned context contains task-discriminative structure while preserving meaningful similarities across tasks.

\subsection{Online Memory Updates}
\label{sec:online_updates}

\begin{figure}[t]
\centering
\begin{subfigure}[t]{0.495\linewidth}
\centering
\includegraphics[width=\linewidth]{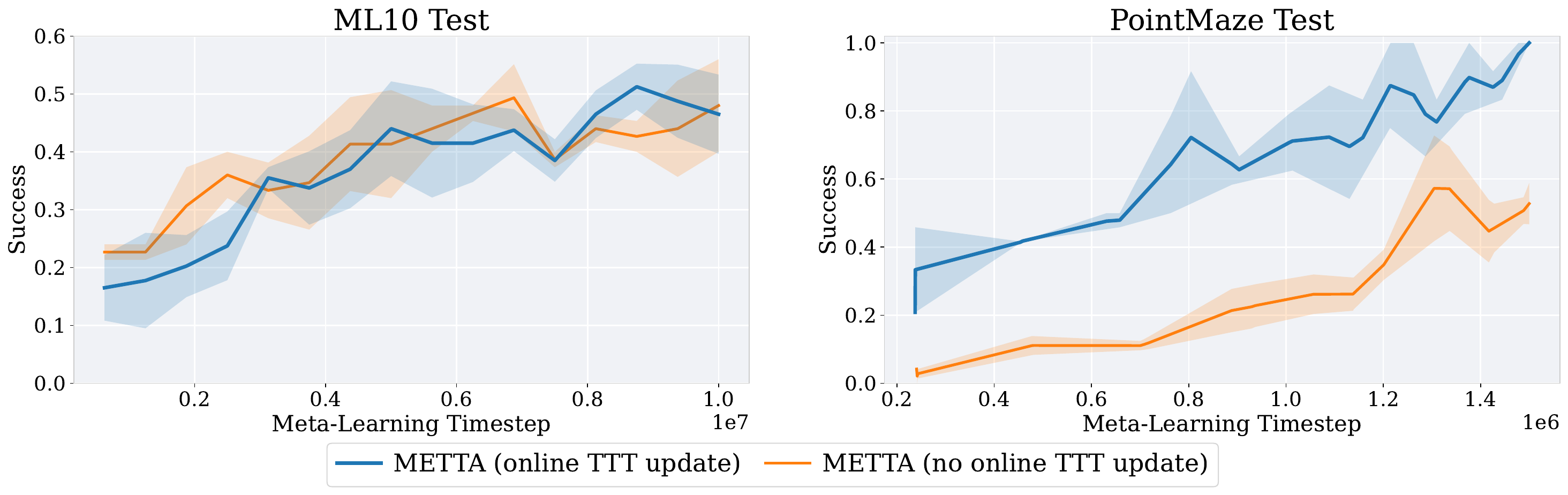}
\caption{ML10 and PointMaze test performance.}
\label{fig:online_updates_tasks}
\end{subfigure}
\hfill
\begin{subfigure}[t]{0.495\linewidth}
\centering
\includegraphics[width=\linewidth]{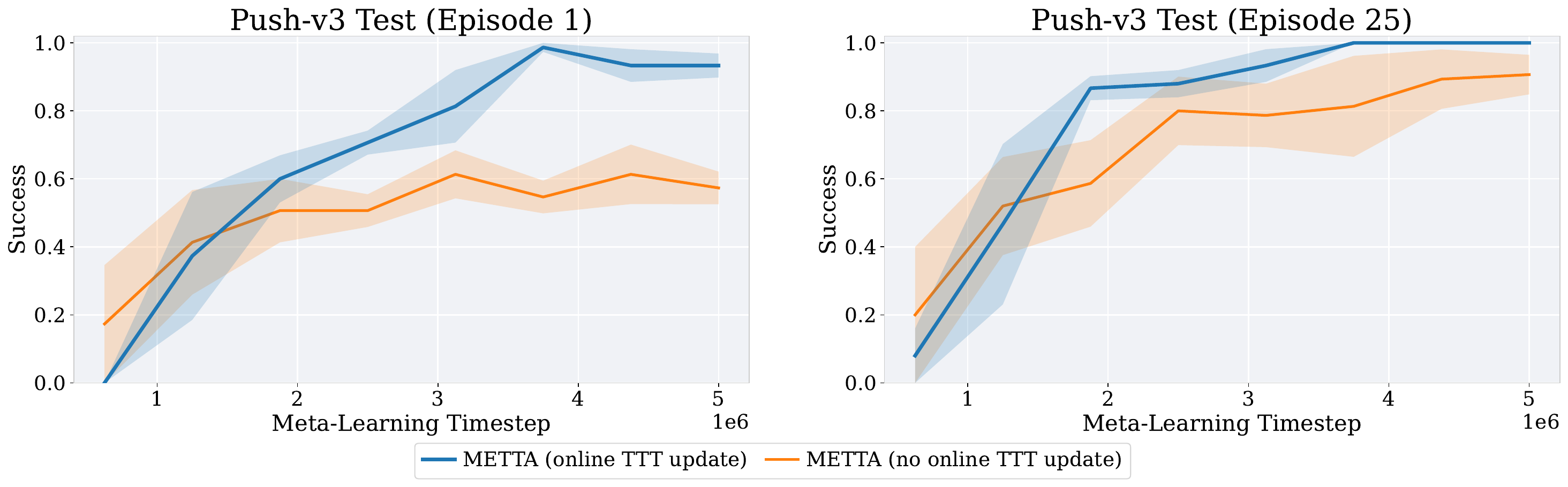}
\caption{ML1 Push adaptation performance.}
\label{fig:online_updates_push}
\end{subfigure}

\caption{Effect of \methodname{} test-time adaptation. The ML1 Push results report adaptation episodes 1 and 25 within each evaluation trial.}
\label{fig:online_updates}
\end{figure}

To isolate the contribution of test-time memory adaptation, we evaluate \methodname{} with its online memory updates disabled while retaining cross-episode aggregation.
Figure~\ref{fig:online_updates_tasks} shows that the effect depends strongly on how task-relevant information is obtained. On PointMaze, disabling the online update causes a large drop in test performance, showing that continual within-episode adaptation is particularly important when useful evidence is sparse and must be accumulated through interaction. In contrast, the two variants perform similarly on ML10, where the goal is observed and adaptation mainly requires identifying the task family, which is easy to distinguish once learned, beyond this, online updating makes little difference.

Motivated by this contrast, we further evaluate the same ablation on ML1 Push, where the goal is hidden and must instead be inferred through interaction. Figure~\ref{fig:online_updates_push} shows that, with online updates, \methodname{} reaches substantially higher success in the first evaluation episode, before any previous-episode information is available for cross-episode aggregation. Without online updates, performance is much lower initially and improves as information accumulates across episodes, but remains below the full model even after 25 episodes. This shows that cross-episode aggregation alone does not recover the benefit of within-episode neural memory adaptation when task-relevant information must be inferred from interaction.

To further examine the effect of the within-episode memory design, we integrate \methodname{} with alternative neural memory architectures, including GRU, Mamba~\citep{gu2023mamba}, and linear Transformer~\citep{katharopoulos2020transformers}, and compare their performance in Appendix~\ref{app:sequence_models}. We also study alternative cross-episode aggregation mechanisms in Appendix~\ref{app:aggregators} to evaluate how the two memory components contribute to adaptation.

\subsection{Training Efficiency}

\begin{wrapfigure}{r}{0.49\textwidth}
\vspace{-22pt}
\centering
\includegraphics[width=0.9\linewidth]{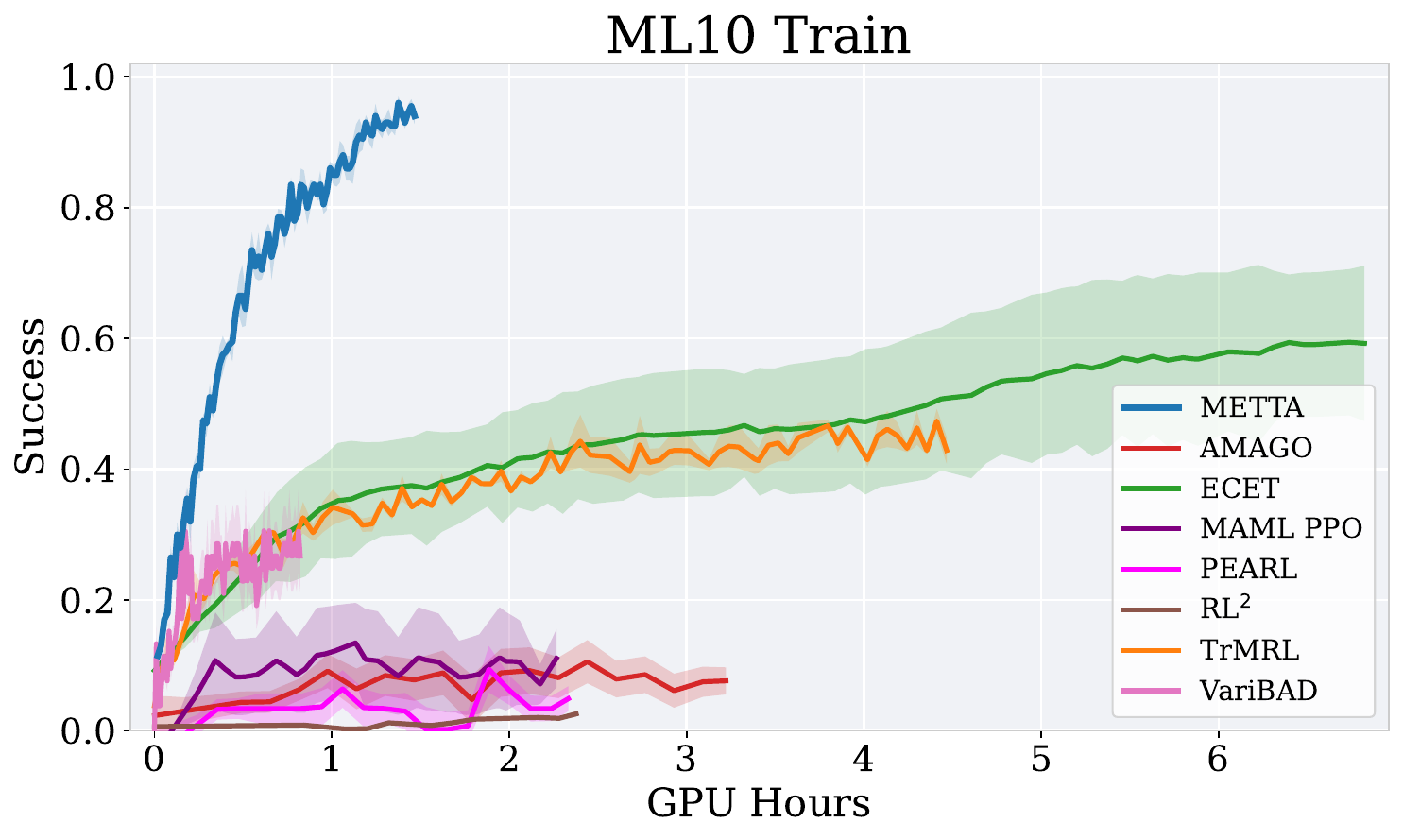}
\vspace{-6pt}
\caption{Success versus GPU hours.}
\label{fig:gpu}
\vspace{-8pt}
\end{wrapfigure}

Figure~\ref{fig:gpu} compares the wall-clock training cost of \methodname{} against transformer-based methods and standard meta-RL baselines on ML10. \methodname{} dominates transformer-based methods, achieving stronger performance with substantially lower GPU-hour cost. At the same time, its training time remains comparable to widely used baselines, while surpassing their performance by a large margin. This shows that neural memory provides a favorable trade-off between long-context adaptation and practical training cost.


\paragraph{Online inference complexity.} Let $T$ denote the episode horizon and $E$ the number of episodes in a trial. Over a complete trial, \methodname{} requires $\mathcal{O}(E^2T)$ computation with attention-based aggregation and $\mathcal{O}(ET)$ with EMA, compared with $\mathcal{O}(E^2T^2)$ for TrMRL~\citep{melo2022transformers} and $\mathcal{O}(ET^3+E^3T)$ for ECET~\citep{shala2025efficient}. Since typically $E \ll T$, the dependence on timestep-level history dominates in long-context settings. TTT incurs a constant per-step cost from its inner update, but this cost does not grow with timestep-level history. Detailed derivations and measured inference time and memory are provided in Appendix~\ref{app:inference_efficiency}.


\section{Conclusion}

We presented \methodname{}, a meta-RL framework based on neural memory and test-time training. The method separates fast adaptation into within-episode online memory updates and cross-episode evidence aggregation, allowing the agent to leverage long interaction histories without attending over the full history. This temporal hierarchy provides an efficient mechanism for long-context adaptation with linear sequence complexity. Unlike standard recurrent approaches that rely on passive memory updates during inference, the TTT module performs gradient-based memory updates, providing an additional mechanism for fast adaptation to previously unseen task families.

Experiments on HalfCheetahVel, Meta-World, and sparse-reward PointMaze showed that \methodname{} improves adaptation across both standard and long-context meta-RL settings while maintaining favorable computational efficiency. 
The gains were especially strong on unseen held-out Meta-World task families, providing further evidence that neural memory with test-time adaptation can support task-family-level OOD generalization to previously unseen, novel task families.  Our findings also suggest a promising direction for future work that extends \methodname{} beyond the fixed-task trial setting considered here, for example by incorporating temporal or recency information into the cross-episode memory for continual or nonstationary environments that require fast online adaptation as task structure, dynamics, or rewards change over time.

\section*{AI Use Statement}
We used generative AI tools to assist in polishing the language, improving clarity and grammar, and debugging the experimental code. These tools did not contribute to the core research ideas, methodological design, or experimental design of this work. All technical decisions, experiments, analyses, and interpretations were carried out and verified by the authors, who take full responsibility for the content of the paper.



\section*{Reproducibility Statement}

The main paper and appendix document the model architecture, training procedure, benchmark settings, hyperparameters,
statistical protocol, baseline configurations, and computational environment used in our experiments. Appendix~\ref{app:implement} provides implementation and optimization details, Appendix~\ref{app:benchmark_details} describes the benchmark configurations, and Appendix~\ref{app:experimental_protocol}
documents the baseline implementations and evaluation protocol. The
supplementary material includes the training and evaluation code and experiment configurations needed to reproduce the reported results. We will make the complete repository publicly available upon acceptance.

\bibliography{ref}
\bibliographystyle{iclr2027_conference}

\clearpage
\appendix

\section{Related Work}
\label{app:rw}

Reinforcement learning methods often achieve strong performance within a fixed environment but generalize poorly across tasks or changing dynamics~\citep{sutton2018reinforcement, packer2018assessing, cobbe19generalization}. This limitation motivated research on meta-learning, where models are trained over a distribution of tasks to learn transferable inductive biases that improve adaptation to unseen settings~\citep{hospedales2020metalearning, bengio1990learning, thrun1998learning}.
Early meta-learning work showed that experience across tasks can improve fast adaptation to unseen settings~\citep{finn2017maml, santoro2016meta, vinyals2016matching}.
Meta-RL extends this idea to sequential decision making, where the agent must infer task structure and adapt behavior using limited interaction data~\citep{duan2016rl2, wang2016learning, finn2017maml}. Unlike conventional RL, which optimizes a single policy for one environment, meta-RL explicitly targets fast adaptation under task variation.

Meta-learning methods are commonly divided into optimization-based, metric-based, and memory-based approaches. Optimization-based methods learn parameters or update rules that support rapid adaptation through gradient-based learning~\citep{finn2017maml, nichol2018firstorder, andrychowicz2016learning}. Metric-based methods instead learn embedding spaces for fast similarity-based generalization~\citep{vinyals2016matching}, while memory-based approaches use recurrent architectures or external memory to encode task information over time~\citep{santoro2016meta}. Although most meta-RL work builds on optimization-based learning~\citep{finn2017maml}, recurrent adaptation and latent memory mechanisms became particularly important because RL environments are sequential and partially observable.

Early meta-RL methods mainly relied on recurrent policies that implicitly infer task identity from trajectories by compressing interaction history into recurrent hidden states. RL$^2$~\citep{duan2016rl2} formulated the problem as learning an RL algorithm inside a recurrent network, where hidden states encode adaptation behavior~\citep{wang2016learning}. Subsequent work improved task inference using probabilistic latent variables and context encoders. VariBAD~\citep{zintgraf2021varibad} modeled meta-RL as approximate Bayesian inference over hidden task variables, while PEARL~\citep{rakelly2019pearl} introduced off-policy probabilistic context adaptation and significantly improved sample efficiency. Other influential methods include ProMP~\citep{rothfuss2018promp}, MAESN~\citep{Gupta2018structured}, and Meta-Q-Learning (MQL)~\citep{fakoor2020mql}, which use recurrent or latent context encoders to improve adaptation efficiency and task inference. Gradient-based meta-RL approaches directly adapted MAML to policy optimization settings~\citep{alshedivat2018continuous_adaptation}, enabling fast policy adaptation using a few gradient steps.

A major challenge in meta-RL is balancing exploration and adaptation under partial observability, particularly when task-relevant evidence is sparse or delayed. In many benchmarks, the underlying task is hidden and must be inferred from sparse interactions. This led to methods that explicitly separate task inference from control~\citep{rakelly2019pearl, zintgraf2021varibad}. Bayesian formulations became especially influential because they model uncertainty over latent task identity and support structured exploration policies. VariBAD~\citep{zintgraf2021varibad} and PEARL~\citep{rakelly2019pearl} showed that latent posterior estimation improves adaptation efficiency and robustness. Related work also studied exploration-driven adaptation, where agents learn information-seeking behaviors that reduce uncertainty during early episodes~\citep{Gupta2018structured, stadie2015incentivizing_exploration}. These formulations closely connect meta-RL with partially observable Markov decision processes and Bayesian reinforcement learning~\citep{guez2014bayesadaptive}, where the agent must maintain and update beliefs over latent task structure.

Attention mechanisms and transformer architectures have been applied to reinforcement learning and meta-reinforcement learning to improve long-range retrieval over interaction histories~\citep{mishra2018snail,parisotto2019stabilizing,chen2021decision}. Recent transformer-based meta-RL methods such as TrMRL~\citep{melo2022transformers}, ECET~\citep{shala2025efficient}, and AMAGO~\citep{grigsby2024amago} use sequence modeling over trajectories or episodes to improve task inference and adaptation. These methods reduce reliance on compressed recurrent states and improve long-context reasoning. However, the high memory usage and quadratic computation cost of self-attention make transformer-based methods less suitable for long-horizon sparse-reward settings.
In contrast, \methodname{} uses neural memory with linear-time recurrent updates, preserving both high expressivity and linear sequence complexity, which makes it scalable to considerably longer horizons.

Linear-time alternatives to quadratic self-attention have been studied through recurrent, linear-attention, and state-space sequence models. State-space and linear recurrent models such as S4~\citep{gu2021s4}, S5~\citep{smith2022s5}, RWKV~\citep{peng2023rwkv}, RetNet~\citep{sun2023retnet}, and Mamba~\citep{gu2023mamba} improve scalability through recurrent memory updates and causal streaming inference. Linear and kernelized attention methods~\citep{katharopoulos2020transformers,choromanski2021rethinking} reinterpret attention as associative recurrent computation with subquadratic complexity.
More recent neural memory approaches extend recurrent sequence models beyond passive hidden-state compression by updating memory parameters during inference. Test-Time Training (TTT) layers~\citep{sun2025ttt} formulate the hidden state as a learnable model optimized online through self-supervised updates, while Titans~\citep{behrouz2025titans} introduces persistent neural memory modules for long-range associative retrieval. These architectures provide efficient long-horizon memory and online adaptation mechanisms that are well suited to meta-reinforcement learning and partially observable environments.

\section{Implementation Details}
\label{app:implement}

\subsection{Sequential PPO Optimization}

Standard PPO implementations often construct minibatches by randomly sampling timesteps from the rollout buffer. For adaptive sequence models, this is inefficient. Evaluating a randomly selected timestep requires reconstructing the neural-memory state by forwarding the trajectory from the beginning of the trial up to that point. Random timestep sampling therefore repeatedly recomputes overlapping prefixes.

To improve training efficiency, we perform PPO optimization sequentially over trajectory segments. Each trial is divided into ordered segments that are processed chronologically, and the PPO loss is applied only to the current segment. In our experiments, each segment corresponds to one episode within the trial. This avoids repeatedly reconstructing arbitrary trajectory prefixes and roughly halves the PPO update time in our implementation.

During rollout collection, the TTT fast weights and cross-episode summaries evolve causally and are used for action selection, while the behavior-policy log probabilities and value estimates are stored. PPO optimization then reconstructs the memory state from the stored interactions rather than reusing the rollout memory states. For a segment in episode $e$, all preceding episodes and the required prefix of episode $e$ are reprocessed using the current slow parameters. After the PPO loss for that segment is backpropagated and the slow parameters are updated, the computation graph is discarded. The causal history required for the next segment is therefore recomputed using the updated parameters, so memory states produced before an optimizer step are never reused by later segments.

Gradients are propagated through the complete reconstructed causal computation for the current segment, including the preceding episode representations used by cross-episode aggregation and the differentiable TTT inner memory updates. Although the PPO loss is evaluated only on the current segment, gradients can flow through reprocessed preceding episodes when their representations contribute to the current context. We do not apply a stop-gradient within this reconstructed PPO computation, including through the fast-weight updates, allowing the outer PPO objective to train the slow TTT parameters through their effect on online adaptation. The computation graph ends at each PPO optimizer step; the required history is reconstructed for the next segment using the updated parameters.

Advantage estimates are computed over the flattened trial trajectory using generalized advantage estimation (GAE),
\begin{equation*}
\hat{A}_t
=
\sum_{l=0}^{\infty}
(\gamma \lambda)^l
\delta_{t+l},
\label{eq:gae}
\end{equation*}
where $\delta_t = r_t + \gamma V_{t+1} - V_t$.
To stabilize optimization across heterogeneous tasks, observations are normalized online and rewards are scaled by running discounted-return statistics. After computing GAE over the flattened trial trajectory, advantages are normalized over valid rollout positions, optionally within each task class when task labels are available.

Policy optimization uses the PPO clipped objective,
\begin{equation}
\mathcal{L}_{\text{PPO}}
=
\mathbb{E}
\left[
\min
\left(
\rho_t\hat{A}_t,
\mathrm{clip}(\rho_t,1-\epsilon,1+\epsilon)\hat{A}_t
\right)
\right],
\label{eq:ppo}
\end{equation}
where
\begin{equation}
\rho_t
=
\exp\left(
\log \pi_\psi(a_t\mid s_t,z_t)
-
\ell_t^{\mathrm{old}}
\right),
\label{eq:ppo_ratio}
\end{equation}
and $\ell_t^{\mathrm{old}}$ is the behavior-policy log probability stored during rollout collection and kept fixed throughout all PPO epochs. Meanwhile, $z_t$ is freshly recomputed using the current model parameters. Gradients therefore propagate through the policy head, encoder, neural memory, and cross-episode aggregator when computing the numerator. An old TTT state or context representation does not need to be reconstructed.

\subsection{More on Test-Time Training}

\head{TTT-MLP.}
Following~\citet{sun2025ttt}, we instantiate the inner-loop model $f_\mathcal{M}$ as a residual MLP memory. Specifically, let $f_{\mathrm{MLP},\mathcal{M}}$ denote a two-layer MLP with parameters $\mathcal{M}$ and a GELU activation. The memory model is defined as
\[
    f_{\mathcal{M}}(u)
    =
    u + \mathrm{LN}\!\left(f_{\mathrm{MLP},\mathcal{M}}(u)\right),
\]
where $\mathrm{LN}$ denotes layer normalization. The residual connection and normalization stabilize the online TTT updates, while the MLP gives the memory more expressive capacity.

\head{Chunk-wise Training.}

The step-wise TTT update is causal but difficult to parallelize, since fully online gradient descent computes each gradient using the memory produced by the previous step. To improve efficiency, TTT uses a mini-batch gradient descent approximation over chunks. Let $\mathcal{C}_m=\{x_{m,1},\ldots,x_{m,C}\}$ denote the $m$-th chunk of length $C$, and let $\mathcal{M}_{m,0}$ be the memory inherited from the previous chunk. Instead of computing the gradient at each intermediate memory state, all gradients inside the chunk are computed with respect to the same initial memory:
\[
    g_{m,c}
    =
    \nabla_{\mathcal{M}}
    \ell
    \left(
        \mathcal{M}_{m,0};
        \mathbf{k}_{m,c}, \mathbf{v}_{m,c}
    \right),
    \qquad c=1,\ldots,C .
\]
The memory after position $c$ is then obtained by accumulating the gradients up to that position,
\[
    \mathcal{M}_{m,c}
    =
    \mathcal{M}_{m,0}
    -
    \sum_{i=1}^{c}
    \eta_{m,i} g_{m,i}.
\]
The final memory $\mathcal{M}_{m,C}$ is passed to the next chunk as $\mathcal{M}_{m+1,0}$, preserving chronological dependence across chunks.

Because all gradients in a chunk are evaluated at $\mathcal{M}_{m,0}$, they become independent once the chunk begins and can be computed in parallel. The cumulative update still preserves causal ordering within the chunk, while the chunk-level formulation substantially improves GPU utilization by replacing many sequential updates with batched operations.

\subsection{\methodname{} Algorithm}

Let $\Theta=\{\theta,\phi,\psi,\nu,\zeta\}$ denote all slow outer-loop parameters of \methodname{}. Here, $\phi$ parameterizes the input encoder $h_{e,t}^{(0)}=\phi(x_{e,t})$, $\psi$ parameterizes the policy head $\pi_\psi$, $\nu$ parameterizes the value function $V_\nu$, and $\zeta$ parameterizes the cross-episode aggregator $\mathrm{Agg}_\zeta$. For attention-based aggregation, $\zeta$ consists of the query, key, and value projections together with the parameters of the final layer normalization. The parameters $\theta$ define the TTT module, including the view projections, inner-loop learning rates, and learned initial fast-weight memories $\{\mathcal{M}_{0}^{(l)}\}_{l=1}^{L}$. PPO updates the slow parameters $\Theta$ in the outer loop. In contrast, the episode-specific memories $\mathcal{M}_{e,t}^{(l)}$ are fast inner-loop states. They are initialized from $\mathcal{M}_{0}^{(l)}$ at the start of each episode, updated online during sequence processing, and discarded after the episode.

\begin{algorithm}[t]
\caption{Training \methodname{} with PPO}
\label{alg:train}
\begin{algorithmic}[1]
\Require Training tasks $p_{\mathrm{train}}(\mathcal{T})$, slow parameters $\Theta=\{\theta,\phi,\psi,\nu,\zeta\}$
\Require Trial length $E$, episode horizon $H$, number of environments $B$
\Require PPO parameters $\gamma,\lambda, \epsilon$, number of PPO epochs $N_{\mathrm{PPO}}$, environment minibatch size $B_{\mathrm{mb}}$

\For{each training update}
    \State Sample tasks $\{\mathcal{T}_b\}_{b=1}^{B}$ from $p_{\mathrm{train}}(\mathcal{T})$
    \State Initialize episode-summary history $\mathcal{H}_b \gets \emptyset$ for each environment $b$
    \State Reset environments and initialize previous action, reward, and done inputs to zero

    \For{$e = 1,\dots,E$}
        \State Initialize within-episode TTT memories $\{\mathcal{M}_{b,0}^{(l)}\}_{l=1}^{L}$ from learned initial memories
        \State Reset previous action, reward, and done inputs to zero

        \For{$t = 0,\dots,H$}
            \State Form the interaction input
            \Statex
            \[
                x_{b,e,t}
                =
                [s_{b,e,t}, a_{b,e,t-1}, r_{b,e,t-1}, d_{b,e,t-1}]
            \]

            \State Embed the input
            \(
                h_{b,e,t}^{(0)}=\phi(x_{b,e,t})
            \)

            \State Process the current episode prefix with the TTT module
            \Statex
            \[
                h_{b,e,t}^{(L)}, \{\mathcal{M}_{b,t}^{(l)}\}_{l=1}^{L}
                =
                \mathrm{TTT}_{\theta}
                \left(
                    h_{b,e,t}^{(0)},
                    \{\mathcal{M}_{b,t-1}^{(l)}\}_{l=1}^{L}
                \right)
            \]

            \State Aggregate previous episode summaries
            \Statex
            \[
                z_{b,e,t}
                =
                \mathrm{Agg}_{\zeta}
                \left(
                    h_{b,e,t}^{(L)}, \mathcal{H}_b
                \right)
            \]

            \State Compute policy and value
            \Statex
            \[
                \mu_{b,e,t}, \sigma_\psi = \pi_\psi(s_{b,e,t},z_{b,e,t}),
                \qquad
                v_{b,e,t} = V_\nu(s_{b,e,t},z_{b,e,t})
            \]

            \State Sample action
            \Statex
            \[
                a_{b,e,t}
                \sim
                \mathcal{N}(\mu_{b,e,t}, \sigma_\psi^2 I)
            \]

            \State Step the environment and store transition data
            \Statex
            \[
                \left(x_{b,e,t}, z_{b,e,t}, a_{b,e,t}, r_{b,e,t}, d_{b,e,t},
                \log \pi_\psi(a_{b,e,t}\mid s_{b,e,t},z_{b,e,t}), v_{b,e,t}\right)
            \]
        \EndFor

        \State Store the final episode representation as an episode summary
        \(
            \bar{h}_{b,e} = h_{b,e,H}^{(L)}
        \)

        \State Update the episode-summary history
        \(
            \mathcal{H}_b \gets \mathcal{H}_b \cup \{\bar{h}_{b,e}\}
        \)

        \If{$e < E$}
            \State Reset the environment under the same task
        \EndIf
    \EndFor

    \State Compute advantages $\hat{A}_{b,e,t}$ and returns $\hat{R}_{b,e,t}$ using GAE over the flattened trial
    \State Partition each trial $b$ into chronological PPO segments $\{\mathcal{G}_{b,m}\}_{m=1}^{M}$

    \For{$j = 1,\dots,N_{\mathrm{PPO}}$}
        \State Partition environment indices $\{1,\dots,B\}$ into minibatches $\{\mathcal{I}_q\}_{q=1}^{Q}$ of size $B_{\mathrm{mb}}$

        \For{$q = 1,\dots,Q$}
            \For{$m = 1,\dots,M$}
                \State Form segment minibatch $\mathcal{G}_{q,m}=\{\mathcal{G}_{b,m}: b\in\mathcal{I}_q\}$
                \State Reprocess $\mathcal{G}_{q,m}$ with the corresponding prefixes and episode-summary histories
                \State Compute $\mathcal{L}_{\mathrm{PPO}}$ and $\mathcal{L}_{V}$ on $\mathcal{G}_{q,m}$
                \State Update $\Theta$ by minimizing
                \Statex
                \[
                    \mathcal{L}_{\mathrm{total}}
                    =
                    -\mathcal{L}_{\mathrm{PPO}}
                    +
                    c_v \mathcal{L}_{V}
                \]
            \EndFor
        \EndFor
    \EndFor
\EndFor
\end{algorithmic}
\end{algorithm}

\begin{algorithm}[t]
\caption{Meta-Test Evaluation of \methodname{}}
\label{alg:test}
\begin{algorithmic}[1]
\Require Test tasks $p_{\mathrm{test}}(\mathcal{T})$, trained slow parameters $\Theta=\{\theta,\phi,\psi,\nu,\zeta\}$
\Require Evaluation trial length $E_{\mathrm{eval}}$, episode horizon $H$, number of environments $B$

\State Freeze slow parameters $\Theta$
\State Sample test tasks $\{\mathcal{T}_b\}_{b=1}^{B}$ from $p_{\mathrm{test}}(\mathcal{T})$
\State Initialize episode-summary history $\mathcal{H}_b \gets \emptyset$ for each environment $b$
\State Reset environments and initialize previous action, reward, and done inputs to zero

\For{$e = 1,\dots,E_{\mathrm{eval}}$}
    \State Initialize within-episode TTT memories $\{\mathcal{M}_{b,0}^{(l)}\}_{l=1}^{L}$ from learned initial memories
    \State Reset previous action, reward, and done inputs to zero

    \For{$t = 0,\dots,H$}
        \State Form the interaction input
        \Statex
        \[
            x_{b,e,t}
            =
            [s_{b,e,t}, a_{b,e,t-1}, r_{b,e,t-1}, d_{b,e,t-1}]
        \]

        \State Embed the input
        \Statex
        \[
            h_{b,e,t}^{(0)}=\phi(x_{b,e,t})
        \]

        \State Process the current episode prefix with the TTT module
        \Statex
        \[
            h_{b,e,t}^{(L)}, \{\mathcal{M}_{b,t}^{(l)}\}_{l=1}^{L}
            =
            \mathrm{TTT}_{\theta}
            \left(
                h_{b,e,t}^{(0)},
                \{\mathcal{M}_{b,t-1}^{(l)}\}_{l=1}^{L}
            \right)
        \]

        \State Aggregate previous episode summaries
        \Statex
        \[
            z_{b,e,t}
            =
            \mathrm{Agg}_{\zeta}
            \left(
                h_{b,e,t}^{(L)}, \mathcal{H}_b
            \right)
        \]

        \State Compute deterministic evaluation action
        \Statex
        \[
            \mu_{b,e,t} = \pi_\psi(s_{b,e,t},z_{b,e,t}),
            \qquad
            a_{b,e,t} = \mu_{b,e,t}
        \]

        \State Step the environment and record return and success
    \EndFor

    \State Store the final episode representation as an episode summary
    \Statex
    \[
        \bar{h}_{b,e} = h_{b,e,H}^{(L)}
    \]

    \State Update the episode-summary history
    \Statex
    \[
        \mathcal{H}_b \gets \mathcal{H}_b \cup \{\bar{h}_{b,e}\}
    \]

    \If{$e < E_{\mathrm{eval}}$}
        \State Reset the environment under the same test task
    \EndIf
\EndFor

\end{algorithmic}
\end{algorithm}

\clearpage

\section{Benchmark Details}
\label{app:benchmark_details}

\subsection{Meta-World}
Meta-World is a popular benchmark in the field of meta-reinforcement learning, focusing on robotic manipulation tasks, which enables evaluation in few-shot adaptation. These tasks vary in objectives, object geometries, and interaction dynamics while sharing a common robot embodiment (Sawyer arm) and action space, enabling structured transfer and meta-learning evaluation. We evaluated our algorithm, \methodname{}, on three families of environments within this benchmark.

\head{ML1.}
In ML1, each trial contains episodes from a single task family (e.g., \textit{push}). Following the standard convention used by the baselines, we mask the goal coordinates from the observation, requiring the agent to infer the goal through interaction across episodes.

\head{ML10 and ML45.}
ML10 and ML45 contain multiple robotic manipulation task families. Within a trial, all episodes belong to the same task family, while task variations such as object and goal configurations may change across episodes. Following the standard convention used by the baselines, goal coordinates are included in the observation for ML10 and ML45, rather than being masked as in ML1. Therefore, the persistent information across episodes primarily concerns the underlying task family and task-specific interaction structure. The same trial construction and observation convention are used for all compared methods.

\begin{figure}[t]
\centering
\includegraphics[width=0.95\linewidth]{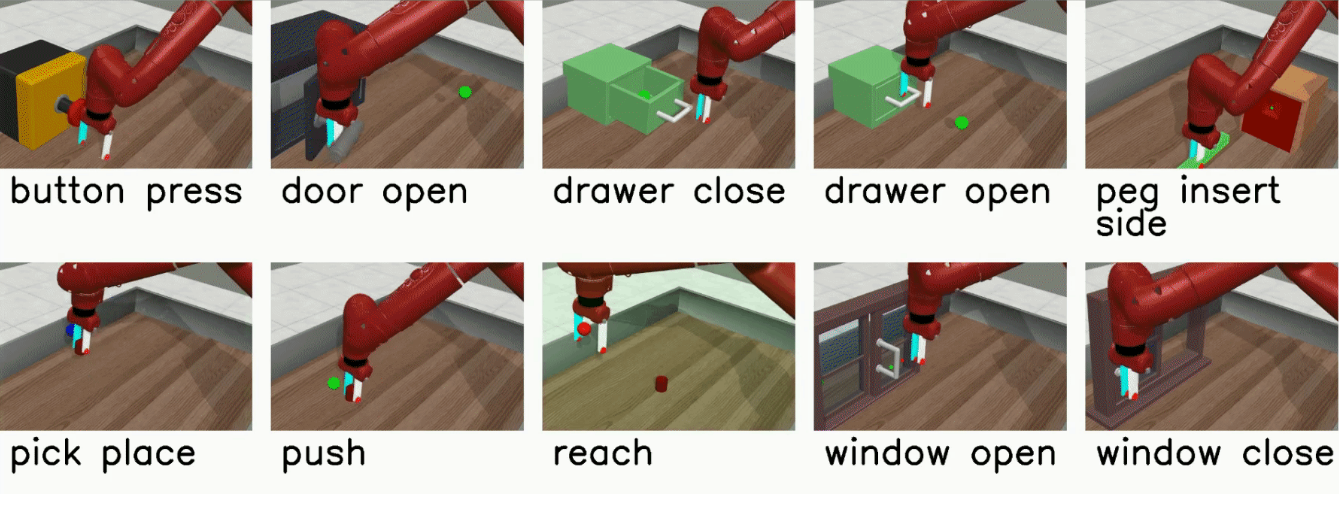}
\caption{
Examples of Meta-World environments illustrate the diversity of robotic manipulation tasks in the benchmark. Figures adapted from the official project page \url{https://meta-world.github.io}. 
}
\label{fig:metaworld_examples}
\end{figure}

\subsection{Mujoco}
We evaluate our method and baselines on a continuous-control task in the MuJoCo simulator, which is widely used in the meta-reinforcement learning literature. Specifically, we consider \textit{HalfCheetahVel}, where the agent must adapt its running speed to a target velocity. In this setup, the target velocity is sampled uniformly from the range $[0, 2]$, and the episode length is set to $100$ timesteps.

\subsection{PointMaze}
We use a PointMaze setup with a $7\times7$ maze, where the agent must navigate to a goal location. The agent and goal are initialized with a minimum shortest-path distance of 10 steps, making successful exploration require long-horizon credit assignment. The environment provides sparse rewards, with reward $+1$ only when the agent reaches the goal, at which point the episode terminates. Within a trial, the task remains fixed across episodes, allowing information acquired during earlier exploration to be reused in subsequent episodes. Training and test tasks differ in their start and goal configurations, with 100 training tasks and 24 held-out test tasks.

\section{Experimental Protocol and Baseline Details}
\label{app:experimental_protocol}

\paragraph{Baseline implementations and hyperparameters.}
For each baseline, we use the implementation and the benchmark-specific configuration recommended by the corresponding repository when available. The baseline implementations are integrated into our codebase through a common training and evaluation pipeline while retaining their default architectures and method-specific settings. Where a repository provides benchmark-specific configurations, we use the
corresponding default configuration. All methods are evaluated under the same environment-interaction budget within each benchmark and using the same task splits and evaluation protocol.
All reported results are computed over five independent random seeds. Learning curves show the mean across seeds, with shaded regions indicating $\pm 1$ standard error of the mean (SEM).

\paragraph{Computational environment.}
All experiments and wall-clock measurements are run on the same server equipped
with an NVIDIA RTX 3090 Ti GPU. All methods are provided with a common hardware and evaluation environment
for the computational comparisons while preserving the implementation choices
specific to each method.

\section{Hyperparameters}

\begin{table}[t!]
\centering
\small
\begin{threeparttable}
\caption{Hyperparameters used in our experiments.}
\label{tab:env_config}
\begin{tabular}{lccc|c|c}
\toprule
\multicolumn{1}{l}{\multirow{2}{*}{\textbf{Parameter}}}
& \multicolumn{3}{c}{\textbf{Meta-World}}
& \multicolumn{1}{c}{\multirow{2}{*}[-0.2em]{\textbf{MuJoCo}}}
& \multicolumn{1}{c}{\multirow{2}{*}[-0.2em]{\textbf{PointMaze}}} \\
\cmidrule(lr){2-4}
\multicolumn{1}{c}{}
& \multicolumn{1}{c}{\textbf{ML1}}
& \multicolumn{1}{c}{\textbf{ML10}}
& \multicolumn{1}{c}{\textbf{ML45}}
& \multicolumn{1}{c}{}
& \multicolumn{1}{c}{} \\
\midrule

\multicolumn{6}{l}{\textbf{Training}} \\
\midrule
\texttt{num\_envs}      & 50 & 50 & 180 & 50 & 25 \\
\texttt{rollout\_steps} & \multicolumn{3}{c|}{500} & 100 & 500 \\
\texttt{trial\_length}  & \multicolumn{3}{c|}{5}   & 5 & 10 \\

\midrule
\multicolumn{6}{l}{\textbf{TTT}} \\
\midrule
\texttt{hidden\_size}        & \multicolumn{5}{c}{128} \\
\texttt{num\_hidden\_layers} & \multicolumn{5}{c}{2} \\
\texttt{mini\_batch\_size}   & \multicolumn{5}{c}{128} \\
\texttt{ttt\_layer\_type}    & \multicolumn{5}{c}{\texttt{mlp}} \\

\midrule
\multicolumn{6}{l}{\textbf{PPO}} \\
\midrule
\texttt{ppo\_epochs}               & \multicolumn{3}{c|}{10} & 10 & 5 \\
\texttt{lr}                        & \multicolumn{5}{c}{3e-4} \\
\texttt{ppo\_envs\_per\_minibatch} & \multicolumn{5}{c}{10} \\
\texttt{policy\_hidden\_sizes}     & \multicolumn{5}{c}{64,64} \\
\texttt{value\_hidden\_sizes}      & \multicolumn{5}{c}{64,64} \\
\texttt{gamma}                     & \multicolumn{5}{c}{0.99} \\
\texttt{gae\_lambda}               & \multicolumn{5}{c}{0.95} \\
\texttt{clip\_epsilon}             & \multicolumn{5}{c}{0.2} \\
\texttt{ent\_coef}                 & \multicolumn{5}{c}{0.0} \\
\texttt{vf\_coef}                  & \multicolumn{5}{c}{0.5} \\
\texttt{max\_std}                  & \multicolumn{5}{c}{1.5} \\
\texttt{min\_std}                  & \multicolumn{3}{c|}{0.5} & 0.1 & 0.5 \\
\texttt{init\_std}                 & \multicolumn{3}{c|}{1.0} & 0.5 & 1.0 \\

\midrule
\multicolumn{6}{l}{\textbf{Meta-Test}} \\
\midrule
\texttt{eval\_trial\_length} & \multicolumn{3}{c|}{25} & 25 & 10 \\

\bottomrule
\end{tabular}
\end{threeparttable}
\end{table}

Table~\ref{tab:env_config} reports the main hyperparameters used in our experiments. The parameter \texttt{num\_envs} corresponds to the number of parallel environments $B$. The parameter \texttt{rollout\_steps} denotes the episode horizon $H$, while \texttt{trial\_length} denotes the number of episodes $E$ in each meta-RL trial used during rollout collection and sequential PPO optimization.

For the TTT module, \texttt{hidden\_size} is the hidden dimension of the TTT representations, and \texttt{num\_hidden\_layers} corresponds to the number of stacked TTT layers $L$. The parameter \texttt{mini\_batch\_size} is the chunk length $C$ used for chunk-wise TTT training. The parameter \texttt{ttt\_layer\_type} specifies the inner-loop memory model $f_{\mathcal{M}}$; in our experiments, we use a two-layer MLP.

For PPO, \texttt{ppo\_epochs} is $N_{\mathrm{PPO}}$, and \texttt{ppo\_envs\_per\_minibatch} denotes $B_{\mathrm{mb}}$, the number of environment trials used in each PPO minibatch before applying chronological segment updates. The policy and value hidden sizes specify the MLP hidden dimensions of the actor and critic heads.

\section{Computational Complexity and Inference Efficiency}
\label{app:inference_efficiency}

Let $T$ denote the number of timesteps per episode, $E$ the number of episodes in a trial, $e$ the current episode index, and $t$ the current timestep. We analyze scaling with interaction-history length while treating hidden dimensions and network depths as constants. Existing complexity analyses for sequence-based meta-RL typically characterize the cost of constructing a context representation at a single decision point rather than the computation accumulated throughout an online trial. For example, ECET~\citep{shala2025efficient} reports $\mathcal{O}(T^2+E^2)$ for its hierarchical attention, corresponding to one context construction when the episode-level transformer evaluations are treated as parallel operations. In online meta-RL, however, action selection occurs at every timestep, so the full-trial complexity naturally accumulates this computation across all timesteps and episodes. We therefore report both the per-decision cost and the total computation over the complete trial.

\paragraph{TrMRL.} TrMRL~\citep{melo2022transformers} applies timestep-level self-attention over the accumulated interaction context. At timestep $t$ of episode $e$, the context contains approximately $(e-1)T+t$ transitions, giving a per-decision cost of $\mathcal{O}((e-1)T+t)$. Summing this cost over the full trial gives
\begin{equation*}
\sum_{\ell=1}^{ET}\mathcal{O}(\ell)=\mathcal{O}(E^2T^2).
\end{equation*}
The stored timestep-level history grows as $\mathcal{O}(ET)$.

\paragraph{ECET.} At episode $e$, ECET has a per-decision computation of $\mathcal{O}(eT^2+e^2)$. Since this computation is performed at each of the $T$ timesteps, the total cost for episode $e$ is
\begin{equation*}
\mathcal{O}(eT^3+e^2T).
\end{equation*}
Summing over all $E$ episodes gives
\begin{equation*}
\sum_{e=1}^{E}\mathcal{O}(eT^3+e^2T)=\mathcal{O}(E^2T^3+E^3T).
\end{equation*}
When the computations across the $e$ episodes are parallelized, the effective per-decision time reduces to $\mathcal{O}(T^2+e^2)$, resulting in $\mathcal{O}(T^3+e^2T)$ per episode and
\begin{equation*}
\sum_{e=1}^{E}\mathcal{O}(T^3+e^2T)=\mathcal{O}(ET^3+E^3T)
\end{equation*}
over the full trial. The stored interaction history scales as $\mathcal{O}(ET)$. Since practical memory usage also depends on the implementation and parallelization strategy, we report measured memory usage below.

\paragraph{\methodname{}.} The within-episode TTT memory has a fixed size with respect to the accumulated interaction history, giving $\mathcal{O}(1)$ history-dependent computation per timestep. With attention-based cross-episode aggregation, the current representation attends only to the $e-1$ previous episode summaries, giving $\mathcal{O}(e)$ additional computation. The computation over episode $e$ is therefore $\mathcal{O}(T+eT)$, and over the full trial
\begin{equation*}
\sum_{e=1}^{E}\mathcal{O}(T+eT)=\mathcal{O}(E^2T).
\end{equation*}
Only episode summaries are retained, requiring $\mathcal{O}(E)$ history-dependent memory. With EMA aggregation, both within-episode and cross-episode memory remain constant with respect to history length, giving $\mathcal{O}(ET)$ computation over the trial and $\mathcal{O}(1)$ history-dependent memory.

\begin{table*}[t]
\centering
\caption{Online inference complexity with respect to episode horizon $T$ and number of episodes $E$. Network dimensions and layer counts are treated as constants.}
\label{tab:complexity}
\small
\begin{tabular}{lccc}
\toprule
Method & Per-decision computation & Full-trial computation & History memory \\
\midrule
TrMRL & $\mathcal{O}((e-1)T+t)$ & $\mathcal{O}(E^2T^2)$ & $\mathcal{O}(ET)$ \\
ECET & $\mathcal{O}(T^2+e^2)$ & $\mathcal{O}(ET^3+E^3T)$ & $\mathcal{O}(ET)$ \\
\methodname{} + Attention & $\mathcal{O}(1+e)$ & $\mathcal{O}(E^2T)$ & $\mathcal{O}(E)$ \\
\methodname{} + EMA & $\mathcal{O}(1)$ & $\mathcal{O}(ET)$ & $\mathcal{O}(1)$ \\
\bottomrule
\end{tabular}
\end{table*}

Since typically $E\ll T$, the dependence on timestep-level history becomes particularly important in long-horizon settings. In addition to the asymptotic analysis, we directly measure per-step inference latency and memory usage for both TTT variants and the transformer-based baselines, providing an empirical comparison of their practical inference overhead.

\begin{table*}[t]
\centering
\caption{Measured per-step inference latency and memory usage as context length increases. TTT-MLP and TTT-Linear include the inner updates performed during inference. Lower values are better ($\downarrow$).}
\label{tab:inference_efficiency}
\resizebox{\linewidth}{!}{%
\begin{tabular}{lcc|cccc|cccc|cccc}
\toprule
& \multicolumn{2}{c|}{\methodname{}}
& \multicolumn{4}{c|}{TrMRL}
& \multicolumn{4}{c|}{ECET ($E=5$)}
& \multicolumn{4}{c}{ECET ($E=25$)} \\
\cmidrule(lr){2-3}
\cmidrule(lr){4-7}
\cmidrule(lr){8-11}
\cmidrule(lr){12-15}
Metric
& TTT-MLP
& TTT-Linear
& $T=5$
& $T=25$
& $T=100$
& $T=500$
& $T=5$
& $T=25$
& $T=100$
& $T=500$
& $T=5$
& $T=25$
& $T=100$
& $T=500$ \\
\midrule
Time (ms/step) $\downarrow$
& 2.159 & 1.616
& 0.551 & 0.639 & 1.061 & 9.223
& 0.761 & 0.991 & 2.119 & 21.713
& 1.035 & 2.151 & 7.599 & 105.403 \\
Memory (MiB) $\downarrow$
& 5.034 & 2.397
& 0.335 & 0.356 & 0.570 & 5.333
& 0.792 & 0.863 & 1.701 & 21.388
& 2.534 & 2.857 & 6.813 & 101.115 \\
\bottomrule
\end{tabular}%
}
\end{table*}

Table~\ref{tab:inference_efficiency} supports the scaling trends in Table~\ref{tab:complexity}. At short context lengths, TrMRL and ECET have lower latency because TTT performs an inner update at every timestep. As $T$ increases, however, their inference time and memory grow substantially, while the TTT cost remains independent of the accumulated timestep-level history. At $T=500$, TTT-MLP requires $2.16$ ms per step, providing $4.3\times$ lower latency than TrMRL and $10.1\times$/$48.8\times$ lower latency than ECET with $E=5/25$. TTT-MLP also uses $5.03$ MiB of memory, over $20\times$ less than ECET with $E=25$. The lighter TTT-Linear variant further reduces the cost to $1.62$ ms per step and $2.40$ MiB, reaching $65.2\times$ lower latency and $42.2\times$ lower memory than ECET with $E=25$ at $T=500$. Overall, TTT has a higher constant cost at short contexts but scales substantially better as the interaction history grows.

\section{Additional Results}
\label{app:additional_res}

\subsection{Alternative Neural Memories}
\label{app:sequence_models}

To examine the flexibility of the within-episode memory component, we integrate \methodname{} with several alternative sequence architectures. Specifically, we replace the TTT memory with a GRU, Mamba~\citep{gu2023mamba}, and a linear Transformer~\citep{katharopoulos2020transformers}, while keeping the cross-episode aggregation mechanism, policy and value networks, and training procedure unchanged.

\begin{figure}[t]
\centering
\includegraphics[width=\linewidth]{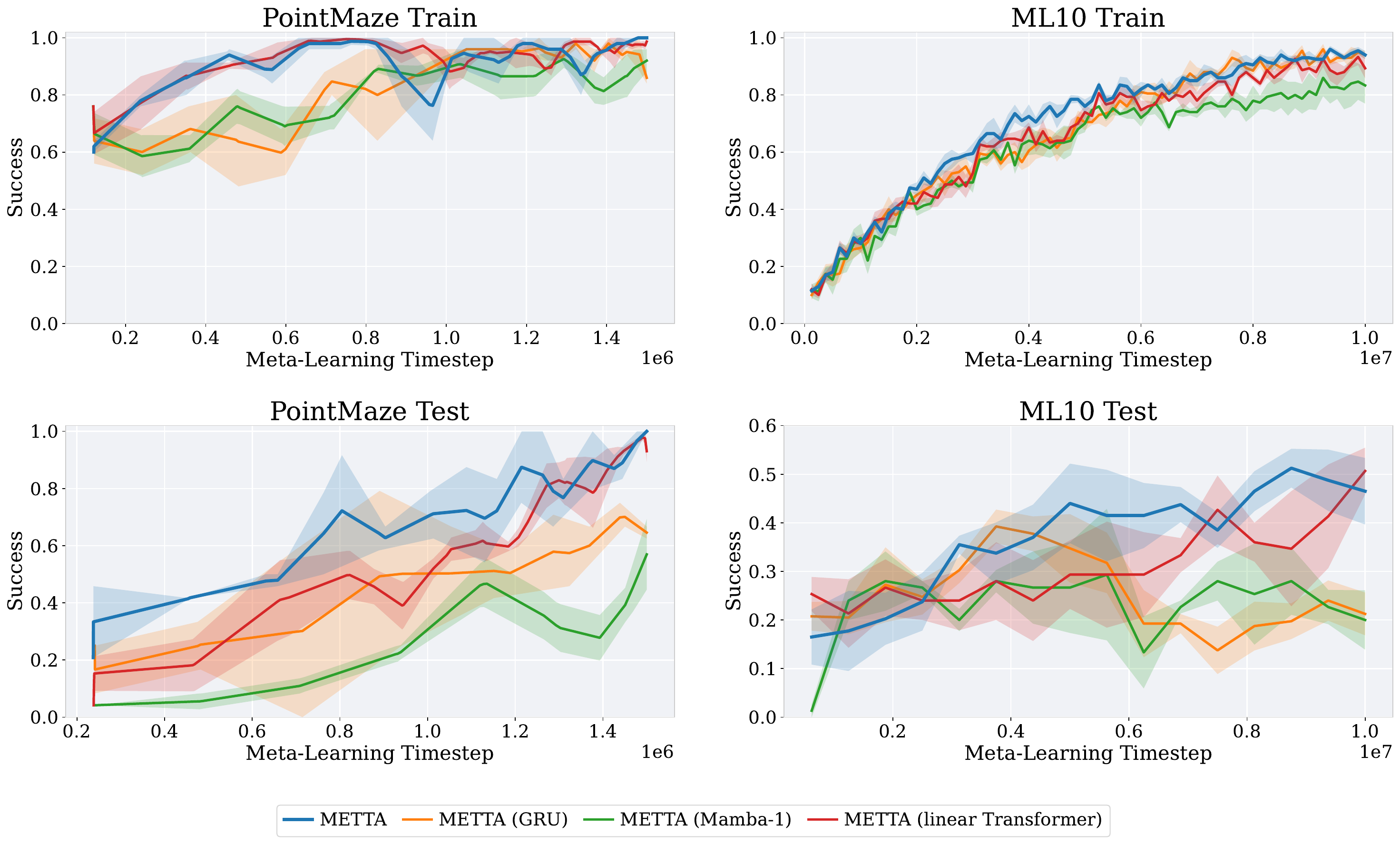}
\caption{Comparison of alternative within-episode neural memory architectures in \methodname{} on ML10 and PointMaze.}
\label{fig:sequence_models}
\end{figure}

Figure~\ref{fig:sequence_models} shows that the \methodname{} framework can operate with different within-episode sequence models, while their adaptation behavior differs substantially. On PointMaze, TTT achieves the strongest and most consistent test performance, while the linear Transformer approaches it later in training and GRU and Mamba remain lower. The differences are also clear on ML10 test tasks: TTT and the linear Transformer achieve substantially stronger generalization than GRU and Mamba, despite all variants learning the training tasks effectively. These results show that the overall framework is not tied to a particular sequence architecture, while the choice of within-episode memory remains important for test-time adaptation and generalization.

\subsection{Cross-Episode Aggregation}
\label{app:aggregators}

We next vary the mechanism used to combine information across episodes while keeping the within-episode neural memory and the rest of \methodname{} unchanged. We compare the default softmax attention with no cross-episode aggregation, an exponential moving average (EMA) with $\alpha=1/2$, and linear attention.

\begin{figure}[t]
\centering
\includegraphics[width=\linewidth]{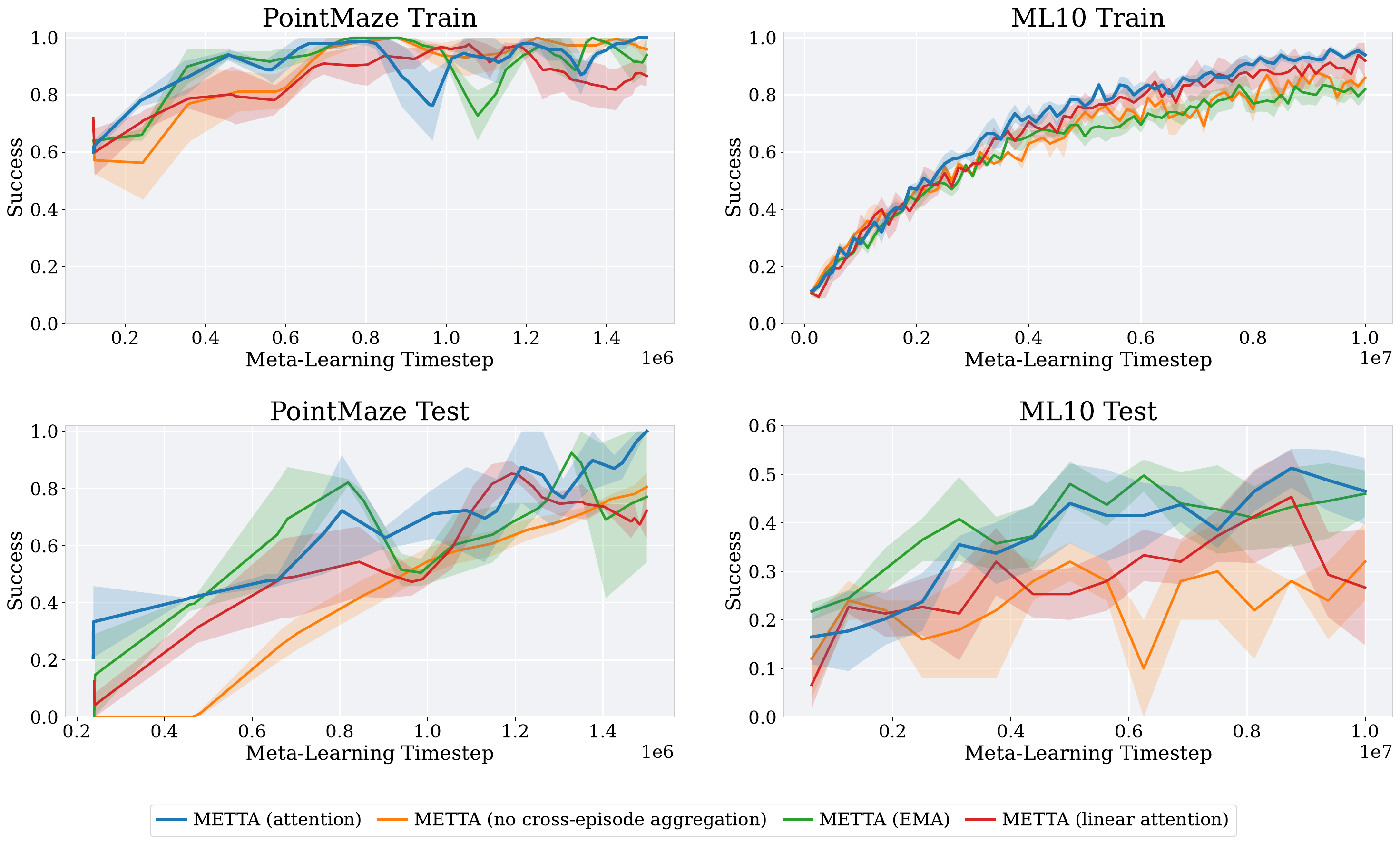}
\caption{Comparison of cross-episode aggregation mechanisms in \methodname{} on ML10 and PointMaze. All variants use the same within-episode neural memory.}
\label{fig:aggregators}
\end{figure}

Figure~\ref{fig:aggregators} shows that the aggregation mechanism has relatively little effect on training performance but becomes substantially more important at test time. On ML10, removing cross-episode aggregation leads to consistently lower test success, remaining around $0.3$, while softmax attention, EMA ($\alpha=1/2$), and linear attention all benefit from reusing information across episodes. On PointMaze, the no-aggregation variant learns much more slowly at test time and remains near zero during approximately the first $0.5$M environment steps, whereas the aggregation-based variants begin improving substantially earlier. This further highlights the importance of reusing evidence collected in previous episodes for adaptation on held-out tasks.

Among the aggregation mechanisms, softmax attention provides the strongest overall trade-off across the four settings. EMA performs competitively on several test settings but is less stable on PointMaze and somewhat weaker on ML10 training tasks. Linear attention closely tracks softmax attention during training and reaches strong PointMaze test performance, but is less consistent on held-out tasks. These results support retaining a selective cross-episode retrieval mechanism while showing that \methodname{} is not restricted to a single aggregation design.

\subsection{Additional Online Memory Update Results}
\label{app:online_updates}

Section~\ref{sec:online_updates} studies the contribution of \methodname{}'s online memory adaptation. Here, we provide the complete training and test curves for this ablation. In all cases, the no-update variant uses the same architecture and cross-episode aggregation as \methodname{}, but disables the online memory updates.

\begin{figure}[t]
\centering
\includegraphics[width=\linewidth]{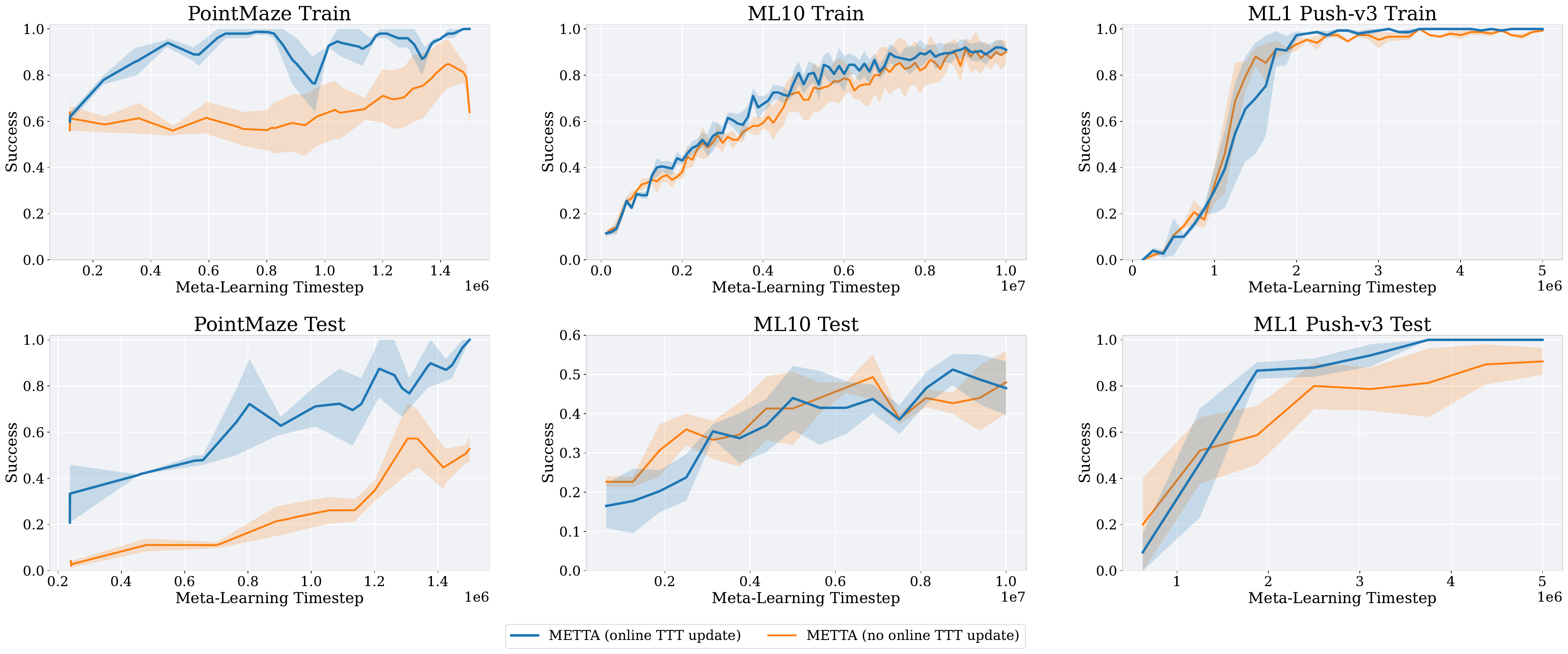}
\caption{Complete online memory-update ablation on PointMaze, ML10, and ML1 Push. Top: training performance. Bottom: test performance.}
\label{fig:online_updates_full}
\end{figure}

\begin{figure}[t]
\centering
\includegraphics[width=\linewidth]{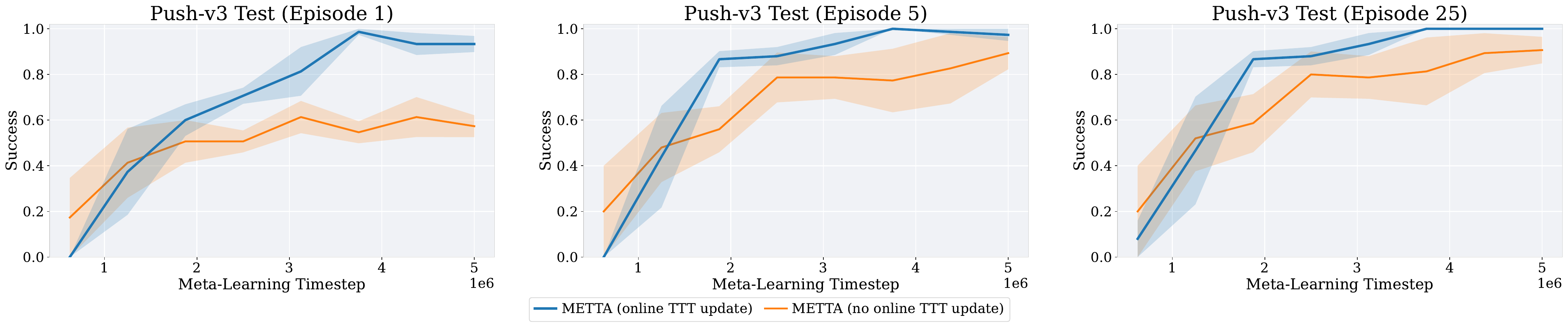}
\caption{ML1 Push test performance at adaptation Episodes 1, 5, and 25 within each evaluation trial. The episode index denotes the progression of adaptation within the same trial.}
\label{fig:online_updates_push_full}
\end{figure}

Figures~\ref{fig:online_updates_full} and~\ref{fig:online_updates_push_full} provide the complete results corresponding to the comparison in Section~\ref{sec:online_updates}. The large difference on PointMaze appears in both training and test performance, with a substantially larger gap on held-out test tasks. On ML10, the two variants remain close throughout training and testing. On ML1 Push, both variants learn the training task effectively, while the online-update model achieves stronger test performance.

\subsection{Memory Capacity}
\label{app:memory_capacity}

We study the effect of the capacity of the within-episode memory by comparing the default TTT-MLP with the lighter TTT-Linear variant. TTT-MLP uses a two-layer MLP as the adaptive neural memory, whereas TTT-Linear uses a single linear layer. Both variants otherwise use the same \methodname{} architecture, cross-episode aggregation, policy and value networks, and training procedure.

\begin{figure}[t]
\centering
\includegraphics[width=\linewidth]{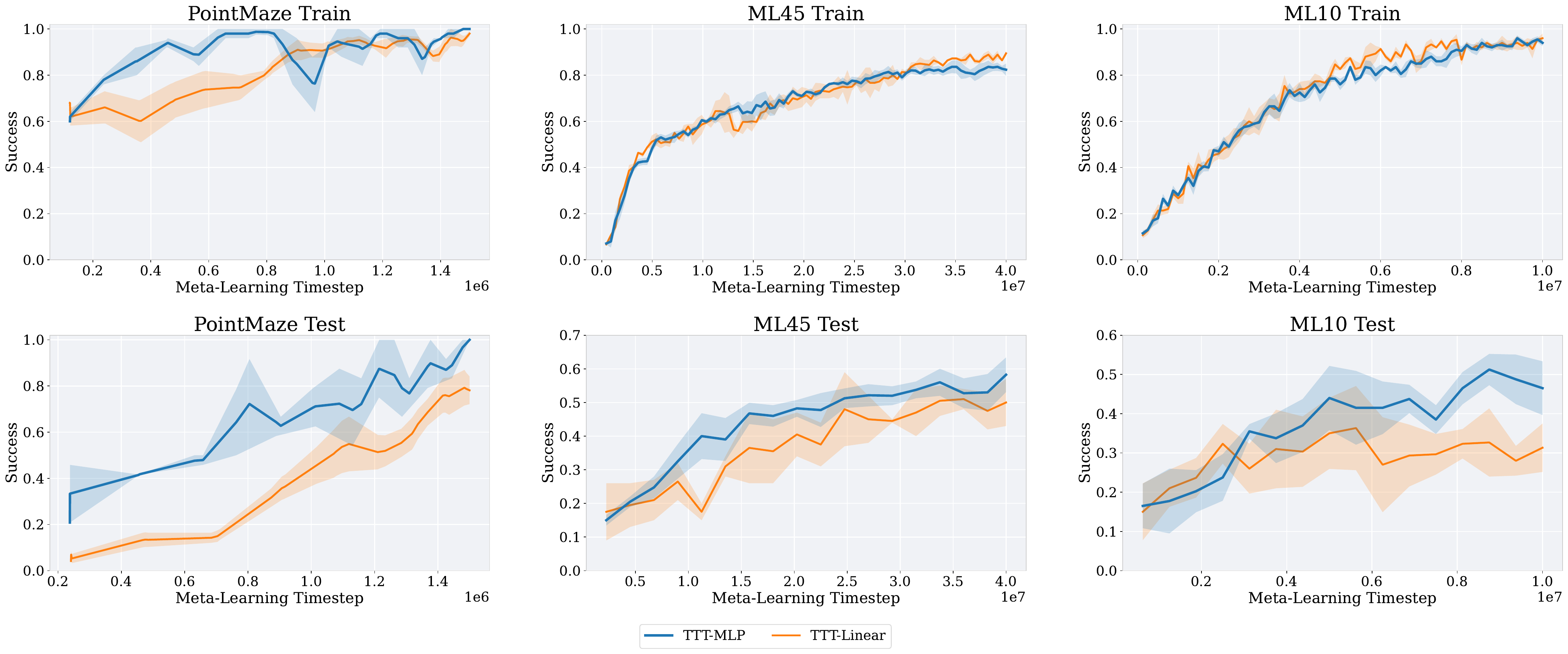}
\caption{Effect of within-episode memory capacity. We compare the default TTT-MLP memory with the lighter TTT-Linear variant on PointMaze, ML45, and ML10.}
\label{fig:ttt_linear_vs_mlp}
\end{figure}

Figure~\ref{fig:ttt_linear_vs_mlp} shows that both memory variants learn the training tasks effectively, while the higher-capacity memory provides stronger test-time adaptation. The difference is largest on PointMaze, where TTT-MLP substantially outperforms TTT-Linear on held-out tasks despite similar training success. TTT-MLP also achieves stronger test performance on ML10 and ML45, although the gap is smaller. These results indicate that increasing the capacity of the adaptive memory is particularly useful when task inference requires integrating richer interaction evidence, while the lighter linear memory remains a viable lower-cost alternative.

\subsection{Separating State and Task Context}
\label{app:state_context}

In the default \methodname{} architecture, the learned context representation $z_{e,t}$ summarizes task-relevant interaction history, while the current state $s_{e,t}$ is provided separately to the actor and critic. We ablate this design by removing the separate state input, forcing the learned representation to jointly encode both the current state and task context.

\begin{figure}[t]
\centering
\includegraphics[width=\linewidth]{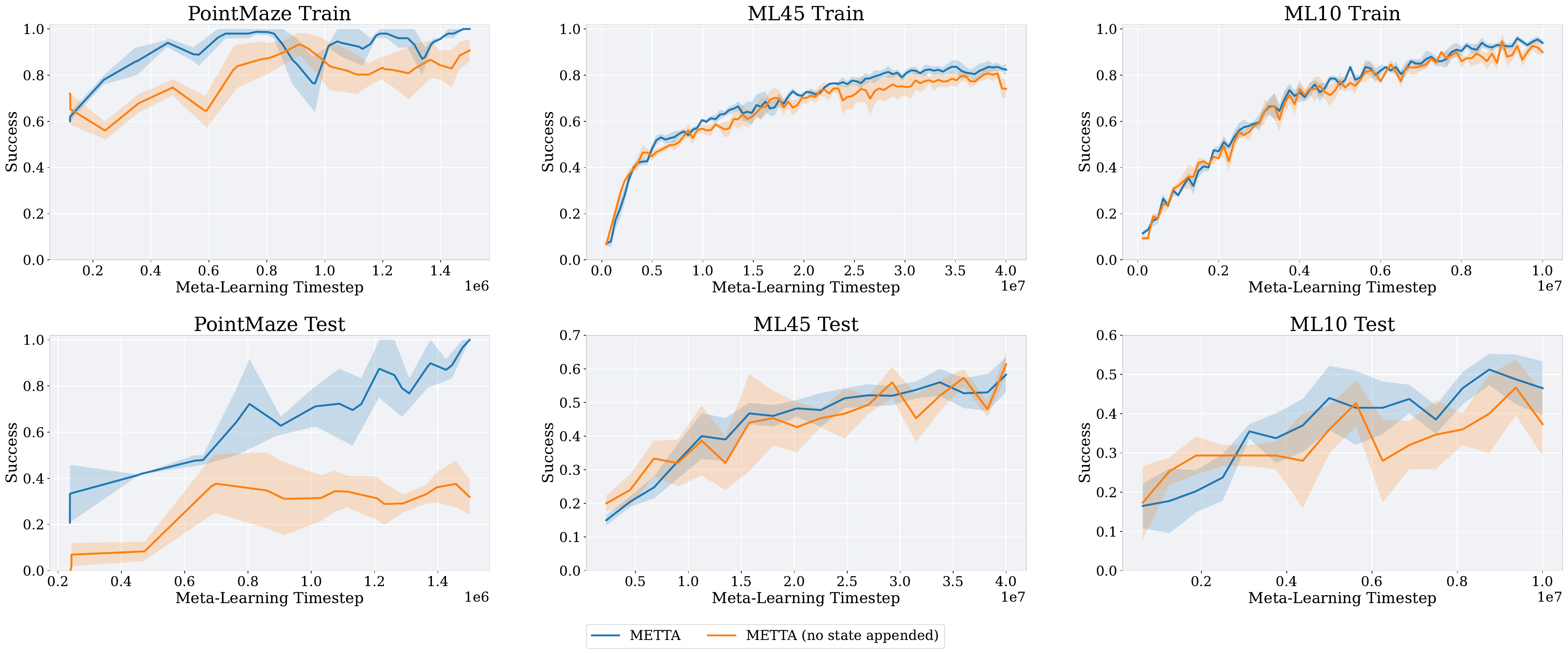}
\caption{Effect of separating the current state from the learned task context. The default \methodname{} provides $s_{e,t}$ separately to the actor and critic, while the ablated variant must represent both state and task information through the learned context.}
\label{fig:state_context}
\end{figure}

Figure~\ref{fig:state_context} shows that explicitly separating state information from task context has relatively little effect on training performance but improves generalization to held-out tasks. The difference is particularly pronounced on PointMaze, where removing the separate state input causes a large reduction in test success despite maintaining strong training performance. The same trend is visible on ML10, while the two variants are closer on ML45. These results suggest that allowing the memory to focus on task-relevant evidence, while providing the instantaneous state directly to the policy and value function, produces a more effective representation for adaptation than requiring a single learned representation to encode both.

\subsection{Qualitative PointMaze Trajectories}
\label{app:qualitative}

\begin{figure}[t]
\centering
\includegraphics[width=\linewidth]{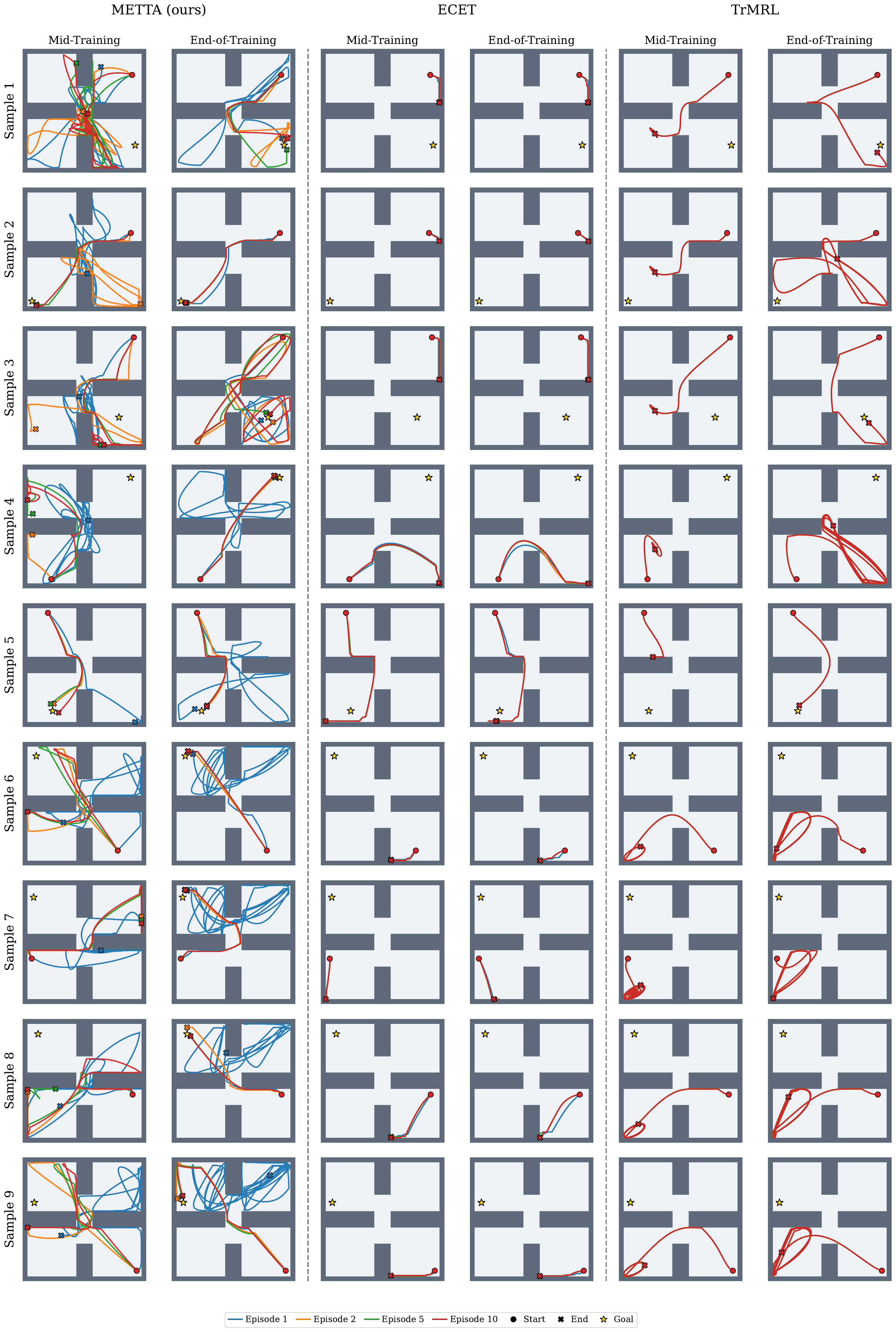}
\caption{Qualitative PointMaze trajectories for \methodname{}, ECET, and TrMRL. Columns correspond to methods, while rows show 9 task instances at mid-training and at the end of training. Colored curves denote Episodes $1$, $2$, $5$, and $10$ from the same trial. ECET and TrMRL frequently revisit similar regions across episodes, whereas \methodname{} covers a wider portion of the maze and progressively produces more goal-directed behavior.}
\label{fig:pointmaze_qualitative}
\end{figure}

Figure~\ref{fig:pointmaze_qualitative} shows PointMaze examples for \methodname{}, ECET, and TrMRL at two stages of learning: halfway through training and at the end of training. For each method, we visualize trajectories from Episodes $1$, $2$, $5$, and $10$ on the same task instances to illustrate how information is used across episodes within a trial.

A clear pattern is that, for ECET and TrMRL, trajectories from different episodes often overlap substantially. This suggests that these agents do not effectively use information gathered in previous episodes, and instead tend to repeat similar exploration patterns across the trial. Even when these methods eventually reach the goal, they frequently revisit previously explored paths rather than refining their search based on earlier failures or successes. In TrMRL, this behavior is consistent with limited effective long-context memory across episodes. For ECET, although it explicitly incorporates cross-episode context, the available context is short, which makes it difficult to use previous episode information effectively in this very low-signal sparse-reward environment. Moreover, this behavior remains visible from mid-training to the end of training, indicating that these methods do not substantially improve their ability to reuse previous episodes over the course of learning.

In contrast, \methodname{} learns a qualitatively different exploration strategy. In particular, it learns to search the maze more broadly in the first episode, and the search space covered by this first episode becomes noticeably larger as training progresses. Later episodes then make better use of the information gathered during this broader initial exploration, leading to more directed behavior toward the goal. These examples, together with the exploration measures in Table~\ref{tab:pointmaze_exploration}, support our claim that episodic neural memory helps preserve and reuse task-relevant evidence across episodes, which is especially important in long-context sparse-reward settings such as PointMaze.

\subsection{PointMaze Exploration Metrics}
\label{app:pointmaze_metrics}

We characterize PointMaze exploration using three measures reported in Table~\ref{tab:pointmaze_exploration}. These measures are computed from the trajectories generated across episodes of each evaluation trial and then averaged across seeds.

\paragraph{Coverage.} Coverage measures the fraction of traversable maze cells visited by the agent before success or episode termination. Higher coverage indicates that the agent explores a larger portion of the available maze rather than repeatedly visiting the same regions.

\paragraph{Episode Jaccard.} For each pair of consecutive episodes, we compute the Jaccard similarity between the sets of maze cells visited in the two episodes after excluding the forced starting cell. For consecutive episodes $e$ and $e+1$ with visited-cell sets $\mathcal{C}_e$ and $\mathcal{C}_{e+1}$, the similarity is
\begin{equation*}
J(\mathcal{C}_e,\mathcal{C}_{e+1})=\frac{|\mathcal{C}_e\cap\mathcal{C}_{e+1}|}{|\mathcal{C}_e\cup\mathcal{C}_{e+1}|}.
\end{equation*}
Lower values indicate that consecutive episodes overlap less and therefore exhibit less repetitive exploration.

\paragraph{Steps.} Steps denotes the number of environment interactions before an episode reaches the goal or terminates at the horizon. When considered together with success performance and maze coverage, fewer steps indicate that broader exploration is achieved without requiring longer trajectories.





\end{document}